%% file: main.tex
\documentclass{ustthesis}

\usepackage{mathpazo,amsmath,amssymb,epsfig,enumerate,bbm,calc,color,ifthen,capt-of} 
\usepackage{algorithm}
\usepackage[noend]{algorithmic}
\usepackage[center]{subfigure}
\usepackage{color,graphicx}

\usepackage{hyperref} 
\usepackage[margin=25mm,textheight=247mm,textwidth=145mm]{geometry}
\usepackage{enumitem}
\usepackage{multirow}
\usepackage{makecell}
\usepackage{esvect}
\usepackage[toc,page]{appendix}
\usepackage{booktabs}
\usepackage{amsmath}
\usepackage{tabularx}

\setlist[description]{leftmargin=\parindent,labelindent=\parindent}

\usepackage[scaled]{helvet} 
\usepackage{courier} 
\usepackage{eulervm} 
\normalfont
\usepackage[T1]{fontenc}

\newcommand{\tab}[1]{\hspace{3mm}}

\title{Finding Good Representations of Emotions for Text Classification}  
\author{Ji Ho~Park}     
\degree{\MPhil}             
\subject{Electronic and Computer Engineering}      
\department{Electronic and Computer Engineering}       
\advisor{Prof. Pascale Fung}     
\depthead{Prof. Bert Shi}    

\defencedate{2018}{08}{01}      

\begin{document}


\maketitle

%



\input{chapter/sec-ack}

\tableofcontents


\listoffigures


\listoftables


\input{chapter/sec-abstract}




\input{chapter/sec-introduction}
\input{chapter/sec-background}
\input{chapter/sec-word-rep}
\input{chapter/sec-sentence-rep}
\input{chapter/sec-abusive}
\input{chapter/sec-conclusion}
\input{chapter/sec-publication}

\bibliographystyle{plain}

\bibliography{ref}



\end{document}

%% file: chapter/sec-ack.tex
\acknowledgments

Firstly I would like to thank my advisor Professor Pascale Fung for leading me into this field and guide me throughout my research. These two years have been a life-changing experience. I have never learned so much in such a short period of time and appreciate all the effort by her to provide me with the necessary support and advice. I am proud to say that I was advised by Pascale and passed her criteria for a research thesis.

I appreciate Professor Bert Shi and Professor Chiew-Lan Tai for taking their times to be in the thesis supervision committee. I hope you enjoy reading my thesis and following my defense presentation. I would also like to thank Professor Yangqiu Song for giving me a good feedback and advice on my research.  

I thank my friend Anik Dey for introducing me to the lab and for being a good mentor and friend throughout my time in HKUST. Also, without the collaboration of Xu Peng, Jay Shin, and Dario Bertero, my thesis would not have been completed. I appreciate all their times discussing and working with me on these works. I feel grateful for all the former and current members of our lab who always support me and openly discussed each other's research. I hope this kind of culture with continue to exist and wish the best in their research as well.

I would like to send my love to parents and brother who are mostly away from me but always give me unconditional support. All the trust and love given by them have been the most important thing in my life. I also want to express my gratitude to my girlfriend Tina Chim because without her next to me for the past 1.5 years it would have been impossible for me to finish my research this well.

Lastly, I want to acknowledge myself for being so persevering and proactive. There were some hard times but I eventually overcame them and achieved many things during the two years. I hope the future self can read these words and be reminded that nothing can beat consistent daily efforts. I hope this thesis is not the end but a start of many more exciting experiences in my life. 

\endacknowledgments

%% file: chapter/sec-abstract.tex
\begin{abstract}
It is important for machines to interpret human emotions properly for better human-machine communications, as emotion is an essential part of human-to-human communications. One aspect of emotion is reflected in the language we use. How to represent emotions in texts is a challenge in natural language processing (NLP). Although continuous vector representations like word2vec have become the new norm for NLP problems, their limitations are that they do not take emotions into consideration and can unintentionally contain bias toward certain identities like different genders.

This thesis focuses on improving existing representations in both word and sentence levels by explicitly taking emotions inside text and model bias into account in their training process. Our improved representations can help to build more robust machine learning models for affect-related text classification like sentiment/emotion analysis and abusive language detection. 

We first propose representations called emotional word vectors (EVEC), which is learned from a convolutional neural network model with an emotion-labeled corpus, which is constructed using hashtags.  Secondly, we extend to learning sentence-level representations by training a bidirectional Long Short-Term Memory model with a huge corpus of texts with the pseudo task of recognizing emojis. We evaluate both representations by performing both qualitative and quantitative analysis and also report high-ranked results in the Semantic Evaluation (SemEval2018) competition. Our results show that, with the representations trained from millions of tweets with weakly supervised labels such as hashtags and emojis, we can solve sentiment/emotion analysis tasks more effectively. 

Lastly, as examples of model bias in representations of existing approaches, we explore a specific problem of automatic detection of abusive language (also known as hate speech). We address the issue of gender bias in various neural network models by conducting experiments to measure and reduce those biases in the representations in order to build more robust classification models.
\end{abstract}

%% file: chapter/sec-introduction.tex
\chapter{Introduction}\label{sec-introduction}
Emotions play an important role in our daily communications. Humans have naturally evolved to express and perceive them in different ways, such as facial expressions, tones of voice, and choice of words. For this reason, developing a sense of empathy toward other people is an essential skill for communicating effectively. Emotions inside language increase the complexity since they not only depend on the semantics but also are inherently subjective, ambiguous, and implicit. For example, the sentence, \textit{''I have not eaten alone for three days,''} merely state a fact that the person consumed food by themselves for a period of time. However, naturally, we can imagine the emotion of the speaker and say, \textit{''oh that must have been pretty lonely,''} and then ask that person to eat together next time. This is called being empathetic, able to understand what the others are feeling and how to correspond to that in a conversation.

Despite the difficulty, accounting for emotions is important in building a machine that truly understands natural language, especially for tasks that are directly related to affect recognition such as sentiment/emotion analysis and abusive language detection, and also those involving human-computer interactions such as dialogue systems and chatbots \cite{fung2015robots}.  As humans can naturally capture and express different emotions in texts,  machines should be able to learn how to infer them as well. 

Some people may think why the world needs empathetic machines that understand human emotions. Popular NLP topics like task-oriented dialogue systems or question and answering, do not seem to be directly related to emotions. However, we believe that machines will take a more active, closer role in supporting humans in the future. Personal assistants like Siri will develop to learn how to make more complex conversations and the expectation of users may grow higher.

To give a more concrete example, many people in healthcare are trying to develop robots that will assist elderly people. This is inevitable due to the growing population of elderly people because of the advancement of medicine and hospitals. Those robots may not only take care of their physical abilities but also their mental states by being a friend and making a conversation. 

When training NLP models, such as chatbots, things do not always go as intended. Famous incident of Microsoft chatbot Tay, which learned directly from users' tweets without any filtering and started becoming racist and spitting out abusive language, gave us a good lesson that a lot of conversation data is not all we need. We also need to be aware of what the machines are learning in terms of empathy. As an extension of this, teaching emotions to machines can include social values and ethics. Understanding what is acceptable to say in the society or when you should be angry.

Representations are the first step to teach machines to understand how humans see the world. In other words, representations are ways of expressing the raw input data in a way that machine learning models can effectively deal with. For example, colored images are converted into two-dimensional matrices with RGB pixel values, because that is how our eye's retina perceive the world visually. Good representations should contain essential information of the data and be a useful input for statistical models to solve problems like classification and regression.

Finding good representations of texts is very challenging since texts are sequences of words which are represented in a discrete space of the vocabulary. The most naive way is to create a dictionary and treat each word as a separate feature inside a sentence. This approach, called the bag-of-words, surprisingly works pretty well for some basic tasks, but do not scale to more complicated natural language processing (NLP) tasks, since it takes away the word order and does not learn much about the relationship among words. 

Many past works have investigated in finding the mapping of words \cite{mikolov2013distributed,pennington2014glove} or sentences \cite{kiros2015skip} to continuous spaces so that each text can be represented by a fixed-size, real-valued vector. Using these vector representations of texts are more effective compared to traditional linguistic features such as word/char n-grams since these vectors have much lower dimensions and encode richer syntactic and semantic information of texts. 

Nevertheless, work focusing on learning representations of emotions inside texts has not been thoroughly explored yet. Previous works mentioned above mostly considered syntax and semantics, which may not be enough for affect-related tasks. Recently, a few works \cite{felbo2017using, tang2014learning} started to show that including sentiment and emotional information in learning representations can be very useful for many relevant tasks. Our work is highly related to these efforts and more literature review will be introduced in Chapter \ref{sec-background}.  

Without a huge corpus, learning robust representations with a powerful generalizable ability is difficult, since language is inherently very diverse, with endless ways to express one's intention and emotion. However, thanks to the endless stream of social media such as Twitter and Facebook, researchers nowadays are lucky enough to have access to almost an unlimited number of texts generated every day. However, annotating these texts with emotion or sentiment human labels is very expensive and difficult. For this reason, a lot of work naturally focused on finding direct or indirect evidence of emotion inside each text, such as hashtags and emoticons \cite{suttles2013distant,wang2012harnessing}, and found them useful to distantly label an emotion of each text. Furthermore, the recent popular culture of using emojis inside social media posts and messages provide us even richer evidence of diverse emotions \cite{felbo2017using, wood2016ruder}. Our work also makes good use of these methods to utilize the streams of data for learning good representations of emotions inside texts.

To learn a good representation from a huge corpus, An appropriate statistical model with sufficient learning capability should be selected. Recently, various deep learning models have been proven to be very powerful for representation learning in the area of natural language, speech recognition, vision, etc. \cite{bengio2013representation}, especially with a huge amount of training data. We explore various deep learning models, such as convolutional neural networks (CNN) and Long Short-Term Memory (LSTM) networks, to learn good representations of emotions of texts. We propose methods for both word and sentence levels since we assume that different levels of representation can capture different information. To prove the effectiveness of these representations, we propose methods to apply them to other relevant text classification problems such as sentiment/emotion analysis and present the performance of our representations. The dataset we use for evaluation includes excerpts of interviews, social media posts (tweets), and online comments.  

Additionally, we broaden the thesis scope by addressing a more specific application, automatic abusive language detection. Abusive language, caused by negative emotions such as anger, fear, and hatred, is an important social issue directly related to our lives. As the number of posts generated on the Internet every day significantly exceeds the capability of human moderators, automatic abusive language detection has become a major demand for many companies such as Google and Facebook. In our work, we apply methods to automatically learn from different levels of representations like characters and words to classify abusive language. Moreover, we discuss the problem of gender bias in the representations of various neural networks learned from existing abusive language datasets and explore ways to reduce those bias to improve the robustness of the representations captured by those models.    

The rest of the thesis is organized as follows:

\begin{description}[labelindent=1.5cm]
\item[$\bullet$] Chapter 2 (Background) provides important reviews in both psychology and natural language processing literature that are fundamentals to our work. 
\item[$\bullet$] Chapter 3 (Emotion Representations in Words) introduces emotional word vectors learned from a hashtag corpus and compare them with other widely used word vectors. 
\item[$\bullet$] Chapter 4 (Emotion Representations in Sentences) presents a method of learning good sentence representations from an emoji cluster corpus and show their effectiveness in sentiment/emotion analysis.  
\item[$\bullet$] Chapter 5 (Abusive Language Detection) discusses our approach of solving automatic abusive language detection from existing public datasets and address the issue of gender bias in the representations of various classification models.
\item[$\bullet$] Chapter 6 (Conclusion) summarizes the results and the significance of learning good representations of emotion in many text classification tasks.  
\end{description}

\newpage

%% file: chapter/sec-background.tex
\chapter{Background}\label{sec-background}

Before proposing our methods, we present some important background knowledge that is fundamental to our work. First of all, we provide some literature review on representations of words and sentences in natural language processing (NLP) research. As mentioned in the introduction, representations are the first things to consider when building a machine learning model, since they fundamentally change how these models can recognize the given input data. We review the existing representation approaches, discuss what are the limitations of them, and connect their relevance to sentiment and emotion.

\section{Categorical Representations}

\begin{figure}[h]
  \centering
  \includegraphics[width=5cm]{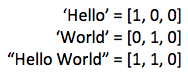}
  \caption{Illustration of `Hello' and `World' as one-hot and ``Hello World'' as bag-of-word. Imagine a 100,000 dimensional vector for a large sized vocabulary.}
  \label{fig:onehot}
\end{figure}

Categorical representations are the most simple way to represent texts. One-hot encodings represent each word in the vocabulary as a binary variable. In a V-dimensional vector (V is the size of the vocabulary), the index of the corresponding word is marked with an integer value 1. All other columns are marked with 0 (Figure \ref{fig:onehot}). Bag-of-word representation is an extension of one-hot. It simply sums up the one-hot representations of words in the sentence. Categorical representations are simple and intuitive, but their limitations are that they do not consider any relational information among words and ignore word orders inside a sentence. Also, in the English language there exists a huge number of words. For this reason, naively representing a word inside a huge vocabulary, such as a one-hot vector of 100,000 words, will easily suffer from the curse of dimensionality. 

\section{Representation Learning in NLP}

In machine learning, feature engineering is still one of the most important processes for building a powerful model. Nevertheless, its limitation is that most of the time those features need to be selected manually, using human knowledge or effort. For this reason, methods to automatically discover the representations needed for downstream tasks like classification have been explored extensively. This process is called ``representation learning'' or ``feature learning.'' 

Representation learning is important because machine learning models often depend on the choice of representations of the input data to perform well \cite{bengio2013representation}. The rise of deep learning models, which have powerful capabilities to learn which features are important, are totally disrupting feature engineering process. Now many tasks, such as image detection and speech recognition, are learned end-to-end fashion, and those models can encode different levels of abstraction inside their layers \cite{Lecun2015}. Robust representations can be learned through either (1) supervised learning (supervised neural networks \cite{Lecun2015} or semi-supervised methods \cite{liang2005semi}, etc.) or (2) unsupervised learning (autoencoders \cite{bengio2013representation}, clustering \cite{coates2011analysis}, etc.).

The area of natural language processing (NLP) is no exception. Especially unsupervised representation learning like pretrained word embeddings have become the most effective method to improve the performance of most downstream tasks by serving as a useful prior knowledge for many machine learning models, effectively substituting categorical representations introduced above. Moreover, traditional NLP feature engineering like syntactic and linguistic n-gram features, Part-of-Speech tags, or sentiment lexicons are being replaced by neural networks and their representation learning capabilities. This thesis is also an extension of those efforts. We will  introduce more background details about word-level and sentence-level representations.

\subsection{Continuous Word Representations}
\begin{figure}[h]
  \centering
  \includegraphics[width=14cm]{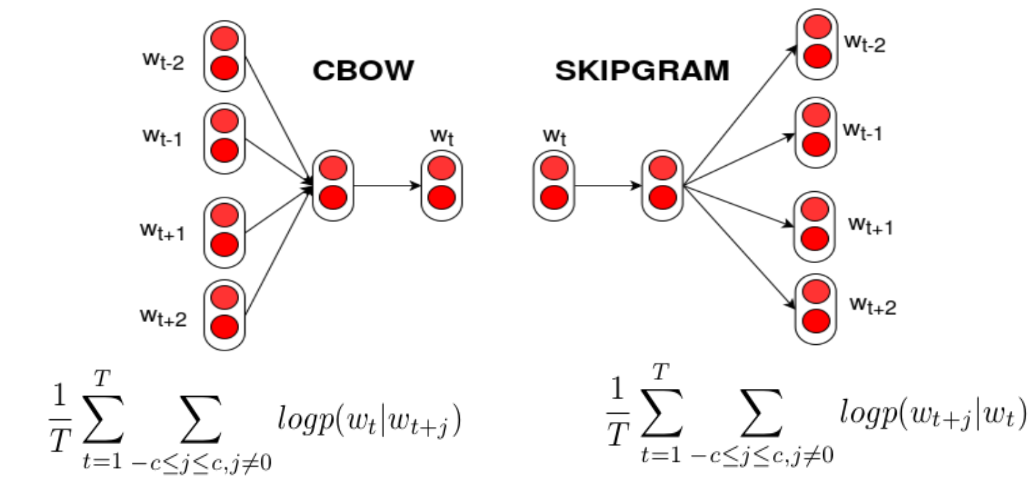}
  \caption{Details of CBOW and Skipgram algorithms to train word2vec embeddings \cite{mikolov2013distributed}}
  \label{fig:word2vec}
\end{figure}

To solve the problematic limitations of one-hot representations, many previous works propose to learn low-dimensional, fixed-sized, real-valued word vectors for every word in the vocabulary, using distributional hypothesis \cite{bojanowski2016enriching, mikolov2013distributed, pennington2014glove}. These vectors are often also called, ``embeddings.'' This hypothesis assumes that a word is defined by its neighboring words (context), and training algorithms like the continuous bag of words (CBOW) or skip-gram (Figure \ref{fig:word2vec}) are proposed to learn vectors that contain syntactic and semantic information of words. These word vectors, or word embeddings, have proven to be useful in many natural language tasks \cite{Collobert2011} when trained with a huge corpus, providing a good prior knowledge of machine learning models. The evaluation of these word embeddings is done by measuring (1) relatedness among words, such as word analogy tasks, (2) coherence among neighboring words, such as word intrusion tasks, or (3) contribution of specific word embeddings to certain downstream tasks, as extrinsic evaluation \cite{schnabel2015evaluation}.

\subsection{Continuous Sentence Representations}

\begin{figure}[h]
  \centering
  \includegraphics[width=\linewidth]{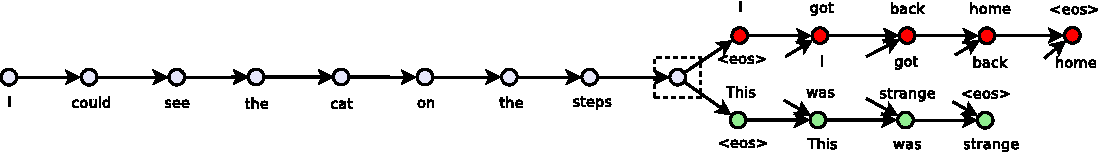}
  \caption{Illustration of how skipthought sentence vectors are trained\cite{kiros2015skip}}
  \label{fig:skipthought}
\end{figure}

Sentences are a sequence of words. Previously, we introduce the most simple way of representing a sentence of using a bag of words representation. This method is easy to implement but loses word order and also has a high dimensionality problem. Another way is to gather all the word vectors of the present words and average them. Such a method is simple and effective but also has a problem of losing word order. A lot of work, inspired by distributed representations of words, proposes methods that effectively learn vector representations of sentences from unlabelled corpora. \cite{kiros2015skip} trains a large RNN encoder-decoder model to encode a sentence and decode the preceding and proceeding sentences (Figure \ref{fig:skipthought}). \cite{le2014distributed} proposes two log-bilinear models that learns a representation sentence, similar to skip-gram algorithm \cite{mikolov2013distributed}. \cite{hill2016learning} proposes a sequential denoising autoencoder model that can be trained without any neighboring sentence. The evaluation of these sentence embeddings can be done by both supervised and unsupervised way \cite{hill2016learning}. Supervised evaluation is to be applied to different sentence classification tasks, such as paraphrase identification, sentiment analysis, etc. Unsupervised evaluation is to measure the related scores between semantically similar sentences.

\section{Sentiment/Emotion in NLP}

In this section, we review the existing works related to sentiment and emotion in the NLP literature. First, we discuss sentiment analysis and its extension, emotion analysis. Furthermore, we discuss recent works that attempt to build word or sentence representations, which are specific to sentiment and emotions in texts. 

\subsection{Sentiment Analysis}
Sentiment analysis is a study of people's attitudes, appraisal, and emotion toward people, objects, topics, and many other entities \cite{liu2012survey}. Due to its practical application like customer review analysis or public opinion mining, it has become one of the most popular fields in natural language processing. There are many kinds of sentiment analysis, but in this section, we focus on document sentiment classification, which is more relevant to this thesis.

The task of document sentiment classification is to find a single opinion on a specific entity of the author. The assumption here is that in each document single author talks about a specific entity with the same opinion throughout the document. Nevertheless, in a real-life situation, such an assumption may not hold. For example, in social media tweet, the author may not directly reveal which entity he or she is talking about. In blog posts, the author may discuss multiple products at once. Such phenomenon makes sentiment classification a challenge. Since most of our corpus used in the experiments are short social media posts like tweets, we keep the same assumption, but we should keep in mind it can fail depending on the source of the data.

Most existing methods for document-level sentiment classification are in supervised learning settings, where a training corpus is given with annotated labels. Labels are usually `positive' and `negative', sometimes containing `neutral'. In some cases, the labels are continuous real-values (e.g. $[0,1]$ negative to positive) or ordinals (e.g. 1 to 5 star reviews). Many previous works \cite{iyyer2015deep, kalchbrennerconvolutional, Kim2014,socher2013recursive, yang2016hierarchical} focus on developing neural networks efficient for these classification tasks, but since our thesis focuses more on representation learning rather than classification models we do not discuss further.

\subsection{Emotion Analysis}

As an extension of sentiment, emotion-related tasks, based on the theory of basic emotions, are also recently proposed. \cite{Mohammad2017,mohammad2018semeval} have introduced annotated datasets and tasks to predict emotion intensities or categories in tweets as shared tasks or competition to the NLP community. We briefly discuss the theory of basic emotions for a better understanding of the tasks. 

\subsubsection{Emotions in Psychological Literatures}

Emotion is defined as ``any conscious experience characterized by intense mental activity and a certain degree of pleasure and displeasure.'' by neuro-scientific and psychological research \cite{damasio1998emotion,ekman1992argument, panksepp2004affective}. It is fundamental to humans since such mental process is necessary to think what is important for survival. For this reason, actions of our body's nerve systems like heart-beating and sweating are deemed relevant to our emotions. 

\begin{figure}[h]
  \centering
  \includegraphics[width=10cm]{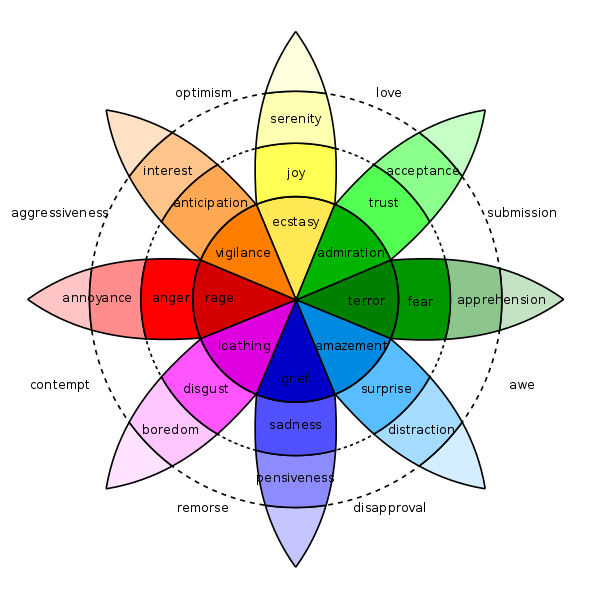}
  \caption{Plutchick's Wheel of emotion}
  \label{fig:wheel-emotion}
\end{figure}

A theory of basic emotions \cite{ekman1992argument,plutchik1984emotions} considers that emotions can be categorized and are universally recognized despite cultural and language differences. \cite{ekman1993facial} shows that facial expressions of different emotions can be easily recognized people from different cultures. Moreover, \cite{ekman1992argument} argues that there are six basic emotions: anger, disgust, fear, joy, sadness, and surprise. \cite{plutchik1984emotions} extends this definition by adding two more basic emotions, trust and anticipation, and further insists that there are opposing pairs of emotions (e.g. joy versus sadness, anger versus fear, trust versus disgust, and surprise versus anticipation). He also proposes the wheel of emotions (Figure \ref{fig:wheel-emotion}), which classifies different emotions on a dimensional basis based on the eight basic categories. There are some works \cite{barrett2006solving,lindquist2015role} challenging such view of emotions being discrete and recognizable. Rather, they claim that emotions are learned and conceptualized through the culture and the language that one grows up with.

Although the on-going debate about the classification of emotion is very interesting, their work put emphasis on the theory of basic emotions since it is widely used in the NLP community \cite{Mohammad2013} and its discrete categorization makes it easy for us to build statistical models on top. Our work depends on the English definition of those basic emotions like joy and sadness because the corpora used in most of our experiments are annotated by humans using these emotion categories. Nevertheless, our proposed methods are not limited to a certain language if there is sufficient textual data in that language.

\subsection{Sentiment/Emotion Specific Representations}

\begin{figure}[h]
  \centering
  \includegraphics[width=6cm]{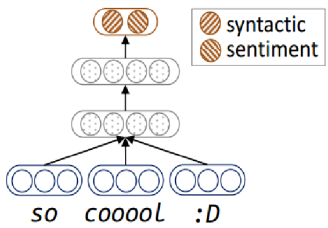}
  \caption{Illustration of Sentiment Specific Word Embedding (SSWE) \cite{tang2014learning}. The word embedding is learned by predicting both the sentiment label and the syntactic context}
  \label{fig:sswe}
\end{figure}

A recent popular approach is to improve the representations of text using sentiment or emotion relevant information. \cite{tang2014learning} proposes to encode the sentiment of words inside embeddings by slightly modifying existing skip-gram algorithm (Figure \ref{fig:sswe}) and also derives sentiment lexicons \cite{tang2014building}. However, this approach is not good since it assumes that all the words in the sentence equally contribute to the sentiment label. \cite{yu2017refining} refines word embeddings for sentiment analysis with sentiment lexicons. \cite{felbo2017using} uses millions of tweets with emojis to model good representations of emotion inside a sentence(tweet) using bidirectional LSTM and demonstrate their significance on many different tasks like document sentiment classification and sarcasm detection. Although their work is very significant, its weakness is that a huge training data (in the magnitude of billions) is needed to train a large neural network model. Also, interpreting their model is not easy, since it does not reveal any word-level information.

Chapter \ref{sec-word-rep} and \ref{sec-sentence-rep} is highly related to the approach of these works focusing on representation learning. Our work tries to be more comprehensive and improve on top of these works. We first focus on learning word-level representations like \cite{tang2014learning} and then extend to sentence-level like \cite{felbo2017using}. In the end, we combine the two outcome representations to solve a shared task proposed in the NLP community. 

\newpage

%% file: chapter/sec-word-rep.tex
\chapter{Emotion representations in words} \label{sec-word-rep}

In this chapter, we present a method to capture emotions inside words and represent this information inside embeddings called emotional word vectors (EVEC). Some words not only have their semantic meanings but also their emotions (valence) inside. For example, when someone says ``I have been eating \textit{alone} for three days," he/she is not merely saying that one has eaten food for three days, but is implying and conveying a message of loneliness. As explained in Chapter \ref{sec-background}, it is a popular and effective practice to represent words as continuous, real-value, fixed dimensional vectors. Words are projected onto points in an N-dimensional space. EVEC are comparable to previous word embeddings in that we use the same vector composition but differ in that we aim to make the vectors of words with similar valence to be clustered together and, therefore, contain better emotional information of words. 

EVEC are learned by using a CNN model trained with a large corpus of tweets, of which emotion labels were obtained by using hashtags as distant labels. By learning representations from a large corpus, our method can capture valences of not only direct emotional words, such as "sad" and "sorrow" but also indirect emotional words, such as "sunshine" and "beach". Such characteristic enables our vectors to substitute traditional emotion lexicons, which are mostly constructed through manual labeling by humans. We visualize some words to interpret what the vectors are capturing by performing Principal Component Analysis (PCA). Moreover, we evaluate the vectors on both word-level and sentence-level tasks and show that they are easy to use and can enhance distributional word embeddings such as Glove or word2vec on different affect-related tasks.

\section{Methodology}

\begin{figure}[t]
  \centering
  \includegraphics[width=15cm]{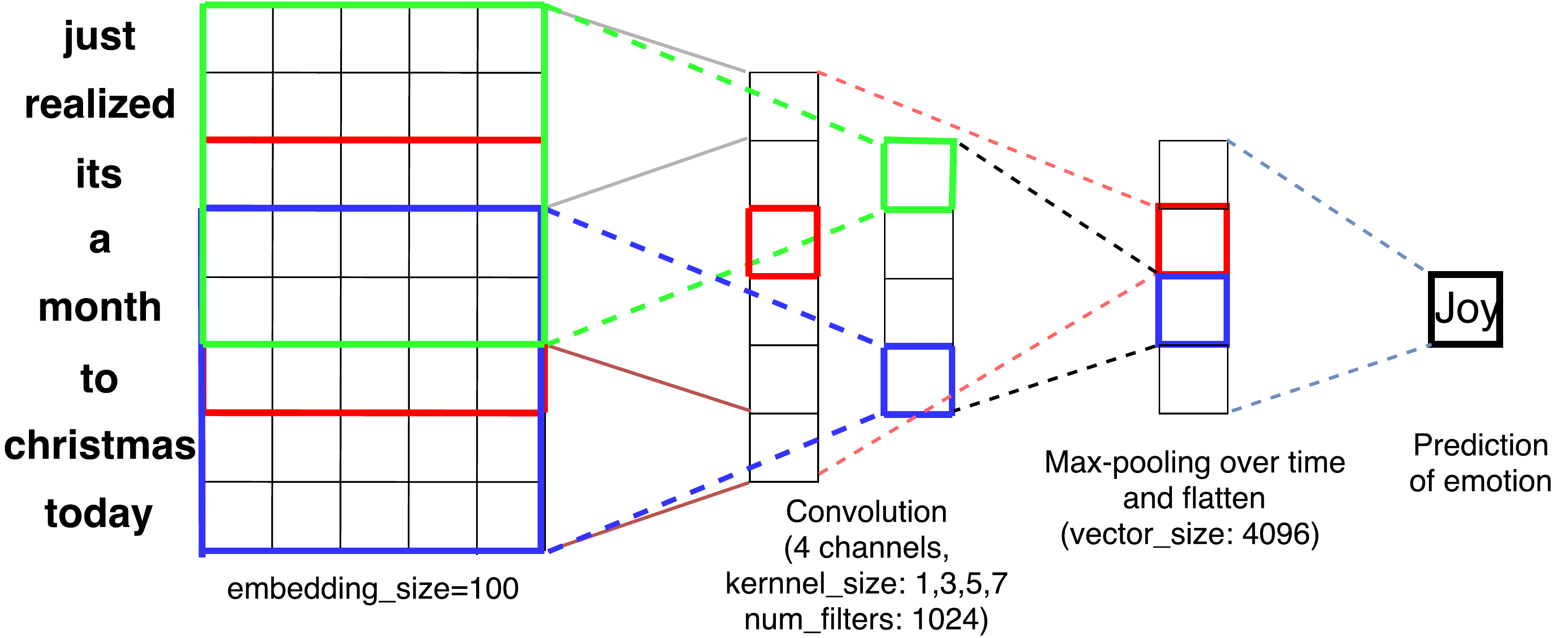}
  \caption{Structure of CNN model to predict the emotion label of a document. Note that the hyperparameters here are merely examples for illustration. They are decided later with a validation set.}
  \label{fig:cnn_architecture}
\end{figure}

\subsection{Model Structure}
Our intuition of learning emotional word embeddings (EVEC) is that given a document with an emotion label, there will be one or more emotionally significant words inside. Nevertheless, machines do not know which specific words are the salient ones. We expect that a deep learning model, which learns representations of data with different level of abstractions~\cite{Lecun2015}, will be able to automatically capture those salient words and encode emotion information in its word embedding layer. For example, if we take the example sentence, "\textit{just realized its a month to Christmas today}", we can easily know that the word \textit{christmas} is the important word for predicting the emotion label "joy". 

For this reason, we choose a CNN structure to learn EVEC (See Figure \ref{fig:cnn_architecture} for example illustration). We assume that the max-pooling function, which only gives weights to important features, will be able to encode more information on the vectors of those emotionally salient words like \textit{Christmas}. We explain the network structure in details below.

Denote each input document as ${[w_{i,1}, w_{i,2}, \cdots, w_{i,M}]}$ where $w_{i,j}$ is the j-th word in the i-th document and M is the number of words. We have an embedding lookup table with size (V, K), where V is the vocabulary size and K is the embedding size. Each word $w_{i,j}$ is transformed into its vector representation from the lookup table. And the input sequence becomes ${[e_{w_{i,1}}, e_{w_{i,2}}, e_{w_{i,3}}, \cdots, e_{w_{i,M}}]}$ denoted as $e_{W_i}$, where $e_{w_{i,j}}$ is the embedding vector for word $w_{i,j}$. For the emotion label of i-th document $l_i$, we use one-hot representation. 

Firstly, 1-D convolution is used to extract local features from the input word vectors.  Specifically, the j-th filter denoted $F_j$ is a dot product between the filter weights and the embedding words in a sliding window with a size $k_j$, giving a feature value $c_{i,j,t}$. We have J filters in total to learn. 
\[c_{i,j,t} = F_j\,* \, e_{w_{i,t:t+k_j-1}} + b_j\]
where $*$ is the 1-D convolution operation. This is similar to looking at the n-gram of the size n=$k_j$ in the sentence.  This is followed by a layer of ReLU activation ~\cite{Nair2010} for non-linearity. After that, we add a max-pooling layer of pooling size $M - F_j + 1$ along the time dimension to force the network to find the most relevant feature for predicting the emotion class correctly. The result of this series of operations is a scalar output of $fm_{i,j}$, which are all concatenated together to produce a condensed vector representation $fm_{i,1:J}$ of the whole input sentence.
\[  fm_{i,j} = \text{Max\_Pooling}\,(ReLU\,(c_{i,j,t}))\]
To classify emotion labels, the condensed vector  $fm_{i,1:J}$ is then projected to target emotion space by adding another fully connected layer with a softmax activation. 

\[  P(l_i | w_{i,1:M} ) = 
 \text{Softmax}(W \cdot [fm_{i,1:J}] + b)\]
We use cross-entropy as the loss function between the predicted emotion distribution and the emotion label. 
\[   loss = - \sum_i  l_i * log P(l_i | w_{i,1:M} )\]

When the model is finished training, we extract the word embedding layer with the size (V,K) as the outcome (EVEC). Each word in the vocabulary will be represented as a K-dimensional vector. The reason that we only use the word embedding layer is that our objective is to obtain the word-level abstraction of emotions from this model. 

\subsection{Dataset}
\subsubsection{Collection}
As mentioned before, collecting a large corpus of emotion-labeled texts is very costly. For this reason, we try to find emotion information from social media posts that can be used as automatic labels. For the dataset used to train EVEC, we used hashtags of Twitter tweets to automatically annotate emotions. This is also known as distant supervision.

Such distant supervision method using hashtags had already been proven to provide reasonably relevant emotion labels by previous works \cite{Wang2012}. Their source of the emotion words came from emotion words list made from \cite{Shaver1987}, a highly cited psychology paper, where \cite{Shaver1987} organized emotions into a hierarchy in which the first layer contains six basic emotions (i.e., love, joy, surprise, anger, sadness and fear), and each emotion has a list of emotion words. \cite{Shaver1987} expanded the list of emotion words by including their lexical variants, e.g., adding "surprising" and "surprised" for "surprise". \cite{Wang2012} further introduced some filtering heuristics, such as only using tweets with emotional hashtags at the end of the tweet to make the distant supervision more relevant. In order to validate this distant supervision method, human evaluation was performed to see whether the distant labels correlated with real human judgement. The result showed that there was high correlation between hashtag labels and human annotation (93.16\%). 

We constructed our hashtag corpus from the data released by \cite{Wang2012} and other public datasets crawled with the same methodology \footnote{http://hci.epfl.ch/sharing-emotion-lexicons-and-data\#emo-hash-data}. We additionally added more tweets between January and October 2017 using the Twitter Firehose API by using the hashtags and filtering heuristics described above to our corpus. The hashtags were transformed to corresponding emotion labels.  A vocabulary with around forty thousand words was created by taking top most frequent words. The total number of documents was about 1.9 million with four classes: joy, sadness, anger, and fear (See Table \ref{table:description-db}). The dataset was split into a train (70\%), validation (15\%), and test set  (15\%) for experiments.

\begin{table*}[h]
\small
\begin{center}
\begin{tabular}{|l|l|l|}
\hline Emotion Label & \% & Samples  \\ 
\hline Joy & 36.5\% & It's been such a great week \bf \#happy \\ 
 Sadness & 33.8\% & I think I miss my boyfriend \bf  \#lonely \\ 
 Anger & 23.5\% & Ignoring me isn't going to make our problems go away. \bf  \#annoyed \\
 Fear & 6\% & What to wear for this job orientation.. \bf  \#nervous \\ \hline

\end{tabular}
\end{center}
\caption{\label{table:description-db} Description of the Twitter hashtag corpus. Hashtags at the end were removed from the document and used as labels. It is hard to construct a well-balanced dataset for all four classes since Twitter users tend to use more hashtags related to happy and sad emotions.}
\end{table*}

\subsubsection{Preprocessing}
When extending the dataset, we only used documents with emotional hashtags at the end, and filtered out any documents with URLs, quotations, or less than five words as \cite{Wang2012} did. Numbers and user mentions were changed to special tokens (e.g. $<number>$, $<user>$), and other non-alphanumeric characters were removed to reduce noisy characters from the corpus. We built our own preprocessing library \footnote{http://www.github.com/jihopark/hltc\_preprocess} based on NLTK toolkit \footnote{http://nltk.org}. This library was reused for most of the data preprocessing of the dataset used in this thesis.

\section{Experimental Setup}
\subsection{Training}

We implemented the CNN model with Keras deep learning library \cite{Chollet2015}. We used Adam \cite{Kingma2015} optimizer and set the learning rate as 0.001. The hyperparameters like number of filters, kernel sizes, and embedding sizes are decided by the emotion category classification performance on the validation set. Embedding sizes can be manually fixed by the experimenter if needed. For our evaluation, we fix the embedding size to 100 or 200 dimension for match the embedding size with other baseline vectors. As a result, we used 4 different sizes of convolution filters with 1024 filters each, 1,3,5 and 7. 

\subsection{Evaluation}

\subsubsection{Clustering emotion lexicons (word-level evaluation)}

In order to measure how much the emotionally similar words are clustered together inside EVEC, we query the nearest neighbors of a word according to the cosine similarity of the word vectors and see how many of these neighbors belong to the same emotion category as the querying word. NRC emotion lexicons \cite{Mohammad2013} are used as the evaluation dataset. They consist of total 3,021 emotional words, and 2,302 words appears at least once in our hashtag corpus. 

We denote each words as $W_1, W_2, \cdots, W_M $, where M is the number of valid words. Each word is labeled with 8 emotions: anger, fear, anticipation, trust, surprise, sadness, joy, and disgust. Here we only keep the words for joy, sadness, anger, and fear, since EVEC are trained for those four emotion categories. 

We first limit the evaluation vocabulary to the NRC Lexicon, because we need the emotion category labels of each word to evaluate. Secondly, for each word $W_m$ in NRC-emotion lexicon, we find the top N-closest neighboring words using cosine distance measure between the word vectors of $W_m$ and all the remaining words in the lexicon. Then, we measure the accuracy of $W_m$ by checking how many of these N neighboring words belong to the same emotion category as $W_m$. Finally, we take the average of the accuracies over all words.

\[ \text{Accuracy} = \frac{\sum_{m=1}^{M} \sum_{n=1}^N 1(l_{W_m} == l_{W_{m,n}})}{N * M}\]
where $l_{W_m}$ means the label of m-th word in NRC-emotion lexicon and $l_{W_{m,n}}$ means the  label of n-th similar word to word $W_m$.

Moreover, we also include results after removing words with low frequency because these words appear too infrequently in our corpus, thus the model has no way to capture enough information about that word. To give a reasonable amount of valid words, we limited the  minimum frequency to 20, which gives 1937 valid words, and to 100, which gives 1081 valid words from the lexicon.

Word2vec vectors trained on the same corpus with the skip-gram algorithm \cite{mikolov2013distributed} is the evaluation baseline. Since skip-gram algorithm does not account any emotional information, EVEC are expected to perform better in this evaluation.

\subsubsection{Affect recognition (sentence-level evaluation)}

An effective representation of words should be able to help to solve related downstream tasks \cite{Collobert2011}. Since the purpose of EVEC is capturing emotional knowledge of word, they should enhance the performance of machine learning models to solve affect-related tasks when used as a input representation. We select four existing sentiment/emotion classification tasks from different data domain and compare EVEC's performance with other word embeddings.

\begin{table}[t]
\begin{center}
\begin{tabular}{|c|cccc|}
\hline \bf{Dataset} & \bf{$N_{sample}$} & \bf{$N_{class}$} & \bf{Labels}  & \bf{Domain}  \\  
\hline  ISEAR & 3,756 & 4 & Emotion (Joy, Sadness, Anger, Fear) & Interview \\
  WASSA & 7,102 & 4 & Emotion (Joy, Sadness, Anger, Fear) & Tweets  \\ 
  SS-Twitter & 2,113 & 2 & Sentiment (Positive, Negative) & Tweets \\
  SS-Youtube & 2,142 & 2 & Sentiment (Positive, Negative) & Comments \\ \hline
\end{tabular}
\end{center}
\caption{\label{table:description-eval-db} Overview of affect-related datasets used for sentence-level evaluation}
\end{table}

\noindent\textbf{Dataset}: Details of the four datasets are following (Table \ref{table:description-eval-db}):
\begin{enumerate}
\item \textit{International Survey on Emotion Antecedents and Reactions (ISEAR)} \cite{Wallbott1986} contains emotion-labeled text excerpts of respondents from a psychological experiment, in which each respondent was asked to report situations on seven major emotions.  We focus on four labels only: joy, sadness, anger, and fear.

\item \textit{Workshop on Computational Approaches to Subjectivity, Sentiment, and Social Media (WASSA) Shared Task} \cite{Mohammad2017} is a Twitter dataset from a shared task for emotion intensity. This dataset is labeled with four emotions and corresponding intensity scores, but we only use the emotion labels for the classification task.

\item \textit{SentiStrength Twitter (SS-Twitter)} \cite{Thelwall2010} is user tweets labeled positive or negative sentiment.

\item \textit{SentiStrength Youtube (SS-Youtube)} \cite{Thelwall2010} is user comments from Youtube labeled positive or negative sentiment.

\end{enumerate}

\noindent\textbf{Comparison with Glove/word2vec}: We want to compare with word embeddings trained with different algorithms and corpora. Two embeddings are used as baseline: Glove \cite{pennington2014glove} trained from a much larger Twitter corpus and word2vec trained on the same corpus as EVEC. 

\begin{table}[t]
\begin{center}
\begin{tabular}{|c|c|c|}
\hline \bf{Embedding} & \bf{Dimension} & \bf{Trained Corpus}  \\  
\hline
Glove    & {100,200} & Twitter Corpus (6B tokens) \\
word2vec & {100,200} & Hashtag Corpus (21.7M tokens) \\
EVEC & {100,200} & Hashtag Corpus (21.7M tokens) \\
\hline
\end{tabular}
\end{center}
\caption{\label{table:description-embeddings} Overview of the embeddings. Note that word2vec and EVEC are trained with the same corpus, but Glove are trained on a much larger corpus.}
\end{table}

\noindent\textbf{Enhancing word2vec/Glove}: Secondly, we want to see whether concatenating existing word embeddings with EVEC also can enhance the performance. Denote one vector of EVEC as $V_e$ and Glove/word2vec vector as $V_0$. Then the concatenation method results in   $[V_e, V_0]$. The assumption is that distributional word representations like Glove or word2vec and EVEC contain different level of information abstraction, so the combination will be stronger and more informative.

To make a fair comparison, we keep the embeddings with same dimension. That is, we compared 200-dim  word2vec/Glove with the concatenation of 100-dim  word2vec/Glove  and 100-dim EVEC to show the benefits of including EVEC. We perform a double cross-validation, 10-fold cross-validation for model selection and averaged the test performance on 10 folds.

\noindent\textbf{Experimental Setup}: To derive representations for classification, we transform all the words into its corresponding vectors. For the out-of-vocabulary words, we use zero vectors. We average all word vectors into one for each sentence. Afterwards, logistic regression classifier is trained with these representations to predict the labels. The reason that we use a simple classifier, rather than a more complex one such as convolutional neural network, is not only that the evaluation datasets are relatively small, but also that we want to focus more on the representational power of the embeddings themselves. We use accuracy for sentiment dataset(SS-Youtube and SS-Twitter) and F1-score for emotion datasets(ISEAR ans WASSA) as evaluation metric. 

\section{Results and Discussion}
\subsection{Results of querying emotion lexicons}

\begin{table}[h]
\begin{center}
\begin{tabular}{|c|l|lll|}
\hline \bf Minimum Frequency & \bf Embedding & \bf top-5 & \bf top-20 & \bf top-100 \\  \hline
\multirow{3}{*}{20} &  word2vec  & 65.0\% & 62.2\% & 59.1\% \\ \cline{2-5}
& \bf EVEC & \bf 67.1\%  & \bf 66.9\% & \bf 65.4\% \\ \hline

\multirow{3}{*}{100}  &  word2vec  & 65.2\% & 63.0\% & 58.6\% \\ \cline{2-5}
 & \bf EVEC & \bf 70.1\%  & \bf 69.6\% & \bf 67.3\% \\ \hline

\end{tabular}
\end{center}
\caption{\label{table:word_comp} Word query accuracy comparison with minimum frequency of 20 and 100, and top-5,20,100 nearest neighbors. EVEC cluster emotion lexicons better than word2vec trained on the same corpus.}
\end{table}

The results in Table \ref{table:word_comp} show EVEC outperform word2vec in clustering emotion lexicon together. Also, we can see that EVEC learn better representations for those words appearing more frequently, since querying with words of minimum frequency 100 works better than those of 20. This result prove that EVEC do put words with similar emotions closer to each other.

\subsection{Results of affect-related tasks}

\subsubsection{Direct comparison with Glove/word2vec}

\begin{table}[h]
\begin{center}
\begin{tabular}{|c|c|ccc|}
\hline  Dataset & Metric & Glove &  word2vec &  EVEC \\  
\hline  ISEAR  & F1 & \bf{.659} & .650 & .645 \\ 
\hline  WASSA & F1 & \bf{.574} & .518 & .573 \\ 
\hline  SS-Twitter & Accuracy & .741 & .755 & \bf{.798} \\ 
\hline  SS-Youtube & Accuracy &  .824 & .830 &  \bf{.840} \\ \hline
\end{tabular}
\end{center}
\caption{\label{table:embedding_result_comp} Comparison between EVEC alone with word2vec/Glove of 100-dim on four datasets. EVEC shows comparable or better performance than other word embeddings.}
\end{table}

EVEC alone show comparable or better than other word embeddings in Table \ref{table:embedding_result_comp}. EVEC have comparable performance with Glove/word2vec. For sentiment datasets, EVEC outperform both of word2vec and Glove. Note that EVEC only captures emotions without syntactic and semantic information and are trained with a much smaller dataset compared to Glove. 

\subsubsection{Enhancing word2vec/Glove}

\begin{table*}[t]
\begin{center}
\begin{tabular}{|l|cccc|}
\hline Dataset & ISEAR & WASSA & SS-Twitter & SS-Youtube \\  
\Xhline{5\arrayrulewidth}
 glove\_200  & .694  & .651  & .742  & .823 \\ 
\hline  glove\_100+\bf{EVEC\_100} &  .719(+3.6\%)  &  .659(+1.2\%)  &  .805(+8.5\%) &  .858(+4.3\%) \\
\Xhline{3\arrayrulewidth}
 word2vec\_200  & .683  & .581  & .769  & .846\\ 
\hline   word2vec\_100+\bf{EVEC\_100} &  .710(+3.9\%)  &   .626(+7.8\%)  &   .806(+4.8\%)  &  .861(+1.7\%) \\
\hline
\end{tabular}
\end{center}
\caption{\label{table:sent_eval} The numbers next to the embedding indicate dimension of the embedding. The percentage inside the parentheses are relative increase from the baseline. This results show that EVEC can help word2vec or Glove to perform affect recognition.} 
\end{table*}

The results in Table \ref{table:sent_eval} illustrate that combining improve the performance, showing that EVEC can enhance distributional representations on classification tasks. For two sentiment datasets SS-Youtube and SS-Twitter, \cite{Thelwall2012} reported separate accuracy for positive and negative classes. Their best model is achieved by logistic regression model using cross validation. We simply take the average and report their results. Accuracy on SS-Twitter and SS-Youtube in \cite{Thelwall2012} is 0.73, 0.59 respectively where our model achieves 0.81 and 0.87.

\subsection{Qualitative Analysis}

\subsubsection{Word Vector Visualization}
\label{word-vector-visualization}

\begin{figure}[p]
  \centering
  \includegraphics[width=\textwidth]{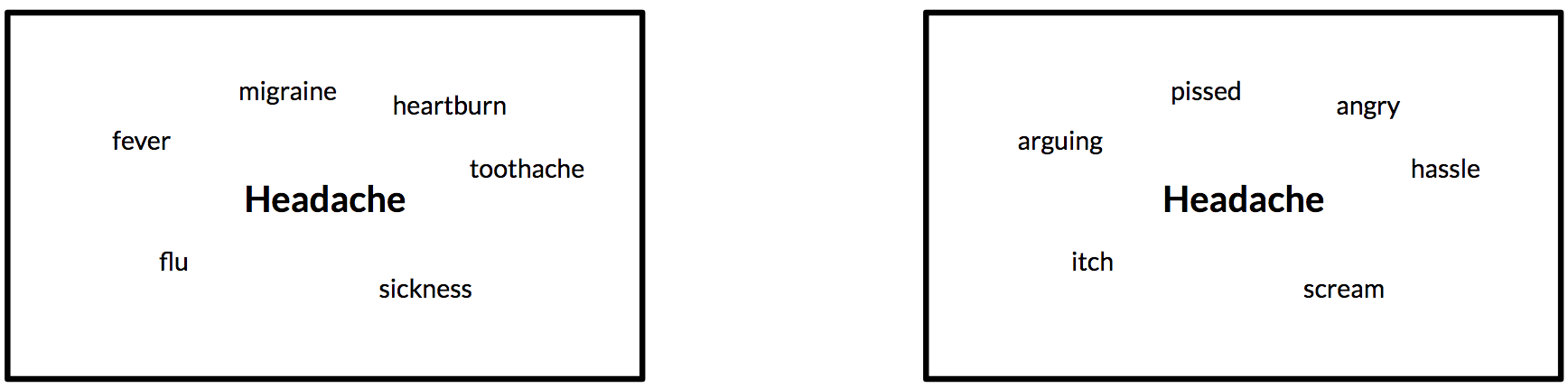}
\caption{Visualization of word ``headache" and its neighbors. Left: (a) Word2vec, Right: (b) EVEC}
\label{fig:headache_combined}

  \vspace*{\floatsep}
  
  \includegraphics[width=\textwidth]{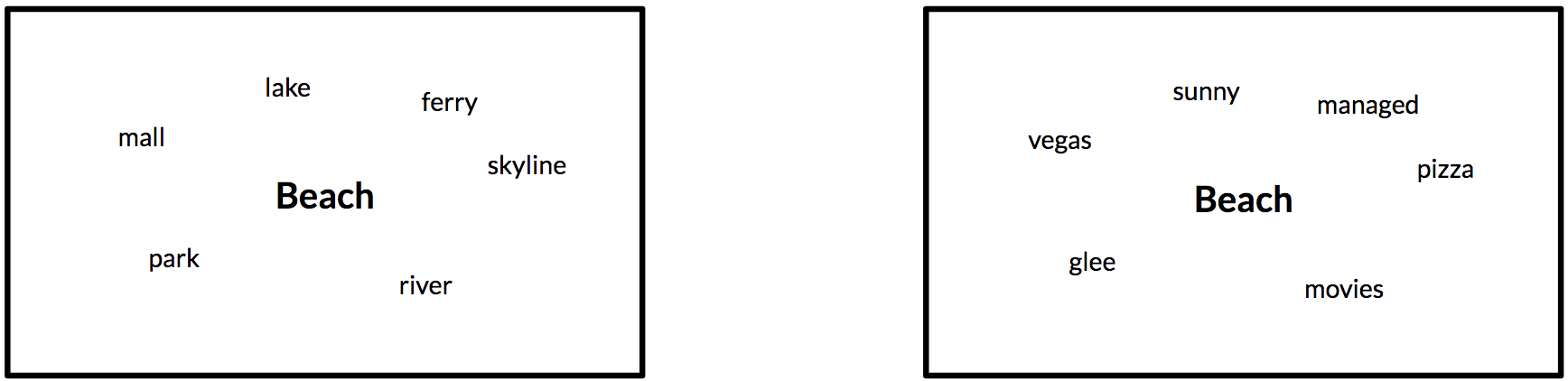}
\caption{Visualization of word ``beach" and its neighbors. Left: (a) Word2vec, Right: (b) EVEC}
\label{fig:beach_combined}

  \vspace*{\floatsep}

\includegraphics[width=\textwidth]{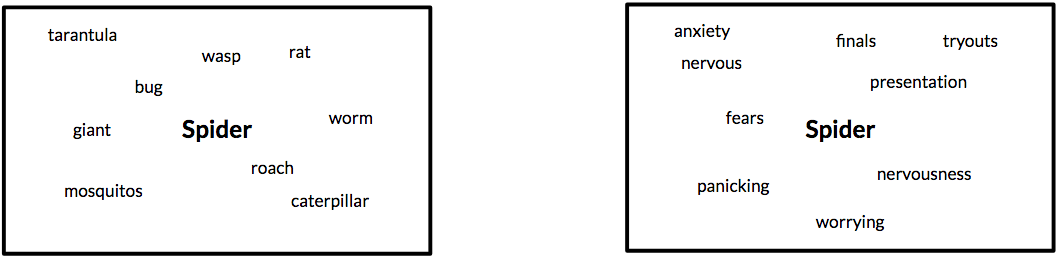}
\caption{Visualization of word ``spider" and its neighbors. Left: (a) Word2vec, Right: (b) EVEC}
\label{fig:spider_combined}

\end{figure}

We projected the trained EVEC into 3D space using Principal Component Analysis (PCA) and visualized each word together with neighboring words. We then compared EVEC with Word2vec vectors trained by skip-gram algorithm on the same corpus. In the Word2vec vector of the word ``headache" cannot capture the emotion of anger. We visualized the EVEC of ``headache" in part (b) of Figure \ref{fig:headache_combined},  where ``headache"  is surrounded by such words as ``pissed", ``anger", ``hassle", and ``scream" that have strong associations with anger.

Figure \ref{fig:beach_combined} and \ref{fig:spider_combined} are also interesting. We can see that EVEC can associate the word ``beach'' with the emotion of joy and the word ``spider'' with the emotion of fear, by looking at their nearest neighbors in the space, whereas Word2vec vectors focus on syntactic and semantic meanings of both words.  

\newpage

\subsubsection{Model Visualization}
Visualizing word vectors is one way to understand what is the outcome of the learning, but visualizing the CNN model is to see what and how we are learning these vectors. Therefore, we performed more analysis on the CNN model used to train EVEC. Inspired by \cite{Li2016} and \cite{Simonyan2013},  we drew a saliency map over the word vectors to show which word is more important in performing the classification. 

To elaborate on how saliency maps work, let's denote a sentence as ${w_1, w_2, w_3, ..., w_M}$, with the emotion label as $l$. We want to see how each word contributes to correctly classifying the emotions. In the CNN model, the mapping from words to prediction of each class is not linear. But we can still use the first-order gradient to approximate the mapping  by Taylor decomposition. 
\begin{align*}
s(w_1, w_2, ..., w_M) &\approx \sum_{i=1}^M k_i * w_i + b \\
k_i &= \frac{\partial s(w_1, w_2, ..., w_M)}{\partial w_i}
\end{align*}
We took the norm of $k_i$, i.e. $|k_i|$ as the saliency score for the word $w_i$. $k_i$ can be  calculated by back-propagating from the output layer to the embedding layer. We  fed the network with a sentence $w_1, w_2, ..., w_M$, and forwarded the network to get the prediction of $ P(l_i | w_{i,1}, \cdots w_{i,M} ) $. Then we back-propagated the gradient to each input embedding dimension.  

\begin{figure*}[t]
  \centering
  \includegraphics[width=\linewidth]{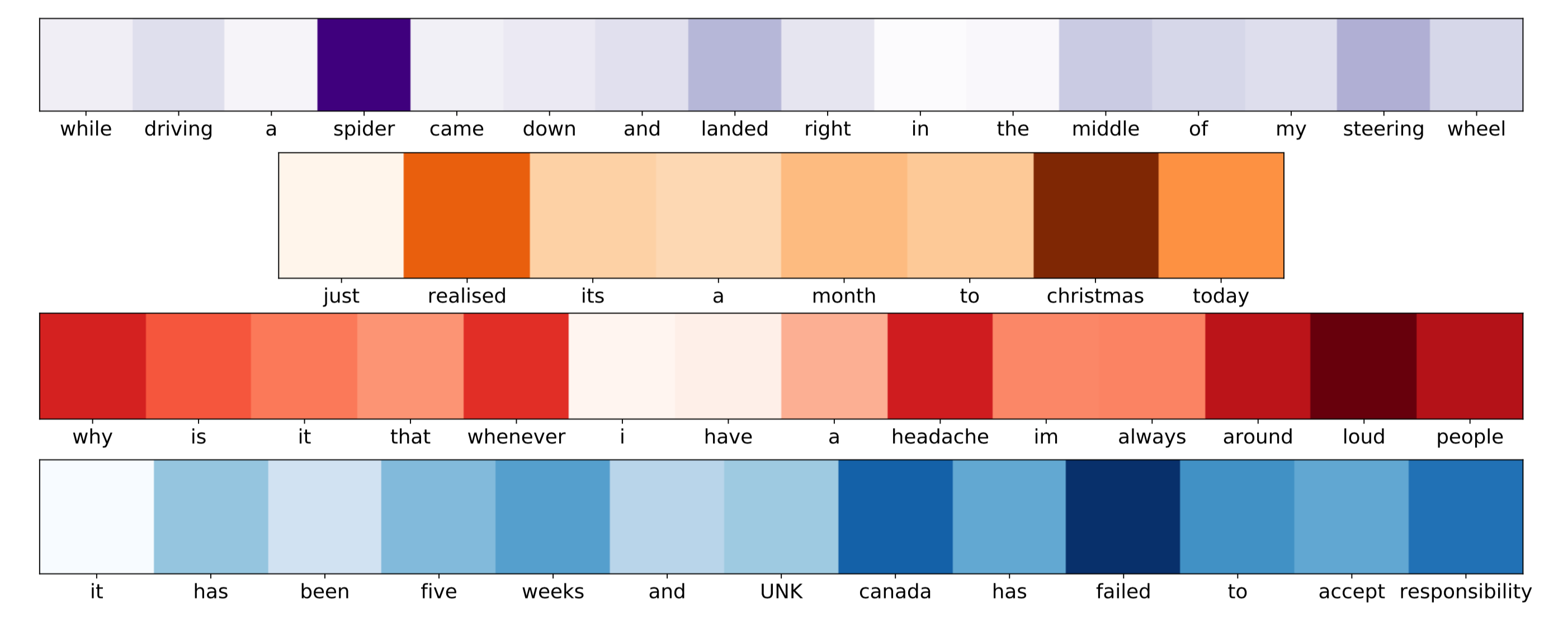}
  \caption{Saliency map of sentence samples from hashtag corpus. From top to bottom, each sentence belongs to the emotion class of fear, joy, anger, and sadness respectively. The darker the color is, the more contributions the word make to the emotion. Emotionally important words in the sentences show more contributions.}
  \label{fig:word_combined}
\end{figure*}

We plotted the saliency score of each word in sentences in Figure \ref{fig:word_combined}. The darker the color is, the more salient the word is. We can see that our model is able to locate emotionally significant words in the sentence,  locating ``spider" for emotion of ``fear" and ``christmas" for emotion of ``joy". ``loud" and ``headache'' for emotion of ``anger", ``failed" for emotion of ``sadness". By looking at this analysis, we can deduce that the model can encode meaningful information of thoese words in their representations, EVEC.

\newpage

\section{Conclusion}

In this chapter we present emotional word vectors(EVEC) that contain emotional semantics into word representation vectors. We propose to learn EVEC by training a CNN model from a Twitter corpus with labels obtained using hashtags. We conduct both word-level and sentence-level evaluations: querying emotional lexicon and performing various affect-related tasks. 

Our results show that EVEC can effectively cluster emotionally similar words together and can be useful input representations for solving affect-related tasks such as sentiment/emotion analysis. The emotional knowledge we extracted from the hashtag corpus can be transferred to other tasks through a simple vector representation of each word. Furthermore, EVEC can be easily used together with other popular word embeddings like Glove and word2vec by concatenating the vectors, and our experiment results show that EVEC can enhance those word embeddings in affect-related tasks. 

Moreover, to better understand how the embeddings are learned, we visualize both our learned vectors and the model architecture, showing that emotional words are recognized in the document, and words with similar valence are clustered together.

Nevertheless, after completing this part of work, another question arose to our mind, ``is word-level information enough to represent emotion?'' Since in our experiments we just average the EVEC of the sentence to form a sentence-level representation, we lose the word order and contexts of words each word. It is natural to think that word-level representations may not be complete to encode emotions of text since not only the meaning but also the emotion of a word when used in a sentence can change depending on the context. EVEC alone is not sufficient to find this out since we only consider the word vectors of the words by themselves. For this reason, the next chapter extends our effort to find good representations of sentences.

\newpage

%% file: chapter/sec-sentence-rep.tex
\chapter{Emotion representations in sentences}\label{sec-sentence-rep}

In this chapter, we explore further by looking into emotion representations in sentences. We present a method to model emotions of sentences into fixed-size, real-valued N-dimensional vectors by using Long Short-Term Memory Networks (LSTM), which can efficiently model sentences as sequences of words. To obtain a more diverse and large collection of emotional texts, we employ emojis in tweets to provide us richer evidence of different emotions. Emojis are proven in previous literature to be highly correlated with the emotions inside their texts and very effective in learning valuable representations for various affect-related tasks \cite{felbo2017using, wood2016ruder}.

We evaluate the emotion sentence representations by applying on tasks of SemEval-2018 Task 1: Affect in Tweets (AIT-2018). We have participated in this competition's five subtasks regarding English tweets on sentiment/emotion intensity prediction (regression/ordinal classification) and multi-label emotion classification. Although these subtasks take different setups, the most important objective is finding a good representation of the tweets regarding emotions. Therefore, they serve as a suitable evaluation of our methodologies proposed so far. We present the performance of the emotion sentence representations both alone and combined with the emotion word vectors (EVEC)  from the last chapter.

\section{Methodology}

\subsection{Model Structure}

Bi-directional LSTM \cite{schuster1997bidirectional} capture the relationship among words inside a sentence by combining two models, forward $\overrightarrow{LSTM}$ and backward $\overleftarrow{LSTM}$. Denote a sentence as sequence of words, ${[x_{1}, x_{2}, \cdots, x_{n}]}$ $n$ is the number of words in the sentence. Forward LSTM considers a sentence from 1 to $n$, whereas backward LSTM does it from $n$ to 1. At the end, we concatenate the hidden state of the two LSTMs and connect to a softmax layer for classification. Below is the formulation of this model.

\[\overrightarrow{h_t} = \overrightarrow{LSTM}(\overrightarrow{W}_{xh}x_t + \overrightarrow{W}_{hh}\overrightarrow{h}_{t-1} + \overrightarrow{b}_{h}) - \text{Input given from timestep 1 to n}\] 
\[\overleftarrow{h_t} = \overleftarrow{LSTM}(\overleftarrow{W}_{xh}x_1 + \overleftarrow{W}_{hh}\overleftarrow{h}_{t-1} + \overleftarrow{b}_{h}) - \text{Input given from timestep n to 1} \] 
\[  P(l | x_{1:N} ) = 
 \text{Softmax}(W \cdot [\overrightarrow{h_t}, \overleftarrow{h_t} ] + b)\]

Note that previous work \cite{felbo2017using} uses a two-layer Bidirectional LSTM model with self-attention, but we decide to use a simpler model due to much smaller dataset size. Our purpose is to replicate the result from previous work with a model with less parameters and smaller data. 

However, for final outcome, we remove the softmax layer after the training and take only the last LSTM hidden states ($[\overrightarrow{h_t}, \overleftarrow{h_t} ]$). This is because our goal is to learn good sentence representations of emotions in sentences, not to learn how to predict what emoji a sentence would include. The emoji prediction task is merely a related, auxiliary task to learn this representation. Therefore, our final model, when given a sentence, can output vector representation of the input through the two LSTMs. We assume that this representation will have rich information of emotions inside, since the model has already learned something about emotions by the emoji prediction task.

\subsection{Dataset}

\begin{figure}[h]
\centering
\includegraphics[width=6cm]{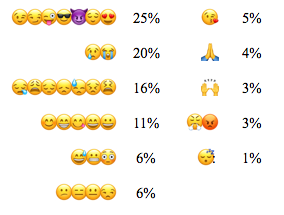}
\caption{11 clusters of emojis used as categorical labels and their distributions in the training set. Because some emojis appear much less frequently than others, we group the 34 emojis into 11 clusters according to the distance on the correlation matrix of the hierarchical clustering from Figure \ref{fig:emoji_hier} and use them as categorical labels}
\label{fig:emoji_dist}
\end{figure}

We crawled 8.1 million tweets with each of which has 34 different facial and hand emojis, assuming these kinds of emojis are more relevant to emotions. Since some emojis appear much less frequently than others, we cluster the 34 emojis into 11 clusters (Figure \ref{fig:emoji_dist}) according to the distance on the correlation matrix of the hierarchical clustering from \cite{felbo2017using}. Samples with emojis in the same cluster are assigned the same categorical label for prediction. Samples with multiple emojis are duplicated  in the training set, whereas in the dev and test set we only use samples with one emoji to avoid confusion.

\begin{figure}[t]
\centering
\includegraphics[width=\textwidth]{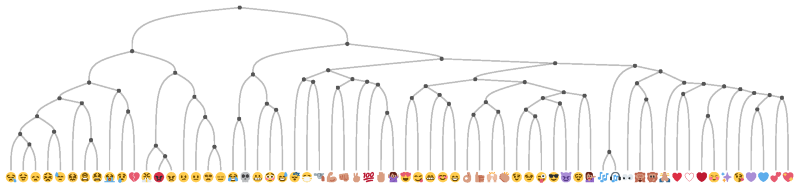}
\caption{Hierarchical clustering results from the predictions of the model of \cite{felbo2017using}. 1.2B tweets were used to train a model to predict the best corresponding emoji out of 64 types.}
\label{fig:emoji_hier}
\end{figure}

\section{Experimental Setup}
\subsection{Training}

The goal of the model is: "given a tweet, predict the most likely emoji cluster." We train a Bi-LSTM model described above with 512 hidden units to predict the emoji cluster of each sample. Cross-entropy loss, a standard for multi-class classification, is optimized by Adam \cite{Kingma2015}. We take part of the dataset to construct a balanced dev set with 15,000 samples per class (total 165,000) for hyperparameter tuning and early stopping. We use 200 dimension Glove vectors pre-trained on a much larger Twitter corpus to initialize the embedding layer.

\subsection{Evaluation}

\subsubsection{Emoji Prediction}

Firstly, we see the outcome of the emoji prediction task. Although the emoji prediction is not our primary goal, we assume that the performance in this task should be correlated to a good representation of emotions. Moreover, we look at the nearest neighbors of some samples from the test set and qualitatively check whether sentences with similar emotions are clustered together. This approach is similar to what we did in Section \ref{word-vector-visualization}.

\subsubsection{SemEval-2018 Task 1: Affect in Tweets}

We want to how well our model can generalize to other datasets. For this reason, we evaluate the emotion sentence representations by applying on tasks of SemEval-2018 Task 1 \cite{mohammad2018semeval} with five subtasks sentiment/emotion intensity prediction (regression/ordinal classification) and multi-label emotion classification. For detailed description of each subtask, please reference \cite{mohammad2018semeval}.

\begin{enumerate}
\item Emotion Intensity Regression (EI-reg): Given a tweet and its emotion category (joy, sadness,fear,anger), predict its emotion intensity score, which is a real-value between 0 to 1.
\item Emotion Intensity Ordinal Classification (EI-oc) Given a tweet and its emotion category (joy,sadness,fear,anger), predict its emotion intensity score, which is a discrete ordinal label (0,1,2,3).
\item Valence (Sentiment) Regression (V-reg): Given a tweet, predict its valence intensity score, which is a real-value between 0 to 1 (negative to positive).
\item Valence (Sentiment) Ordinal Classification (V-oc): Given a tweet, predict its valence intensity score, which is a discrete ordinal label (-3,-2,-1,0,1,2,3)
\item Emotion Classification (E-c): Given a tweet, predict its emotion categories (multi-label among 11 categories)
\end{enumerate}

\begin{table}[t]
\footnotesize
\begin{center}
\begin{tabular}{|c|cccc|c|c|}
\hline \bf{Subtasks} & \multicolumn{4}{c|}{\bf{1/2}} & \bf{3/4} & \bf{5} \\
\hline Category & Anger & Fear & Joy & Sadness & Sentiment & Emotions \\
\hline Train & 1,701 & 2,252 & 1,616 & 1,533 & 1,181 & 6,838 \\
Dev & 388 & 389 & 290 & 397 & 449 & 886 \\
Test & 1,002 & 986 & 1,105 & 975 &  937 & 3,259 \\
\hline
\end{tabular}
\end{center}
\caption{\label{table:dataset} Statistics of the competition dataset for all 5 subtasks}
\end{table}

\begin{table}[h]
\footnotesize
\begin{center}
\begin{tabular}{cccccccc}
\hline \bf{\# of labels} & 0 & 1 &2 &3 &4 &5 &6 \\
\hline \bf{\%} &2.9 &14.3 &40.6 &30.9 &9.6 &1.4 &0.2 \\
\hline
\end{tabular}
\end{center}
\caption{\label{table:label_dist} Number of multi-labels for subtask 5. Most samples have from 1-3 labels, but can have no labels or up to 6 labels.}
\end{table}

SemEval-2018 competition dataset is created by human annotators through crowd-sourcing methods \cite{LREC18-TweetEmo}. Total three datasets are given: emotion intensity (with four emotion categories; Subtask 1/2), sentiment intensity (subtask 3/4), and multi-label emotion classification (subtask 5). 

For emotion and sentiment intensity datasets, each tweet sample has both an \textit{ordinal label} (coarse; \{0,1,2,3\} for emotion, \{-3,-2,-1,0,1,2,3\} for sentiment) and real-value \textit{regression label} (fine-grained; [0,1]). For multi-label emotion classification dataset, each can have none or up to six number of multi-labels (Table \ref{table:label_dist}). 

Since the competition dataset is relatively small, having around 1-2,000 tweets per emotion category, learning good representations of emotions from external source is crucial for boosting the task performance. We use the trained Bi-LSTM model as a feature extractor by passing all tweets in the competition dataset through the model and employ the resulting vector representations as inputs for traditional machine learning models. 

For regression and ordinal classification tasks, we simply use Support Vector Regression (SVR) and Kernel Ridge Linear Regression. Ordinal classification is done using the best regression model by finding the mapping between the regression and ordinal labels. 

For multi-label classification, we use a regularized linear regression and ensemble of classifier chain. Again, the reason we use very simple models is to see the power of our learned representations.

Furthermore, we also report the results on the competition when our emoji sentence representations are used together with EVEC from previous chapter and tweet-specific features, such as number of upper cases and exclamation marks, which are removed from our preprocessing steps. This is to prove that sentence-level and word-level representations can be combined together to be much robust and informative. So the features we used for our experiments are as follows:

{\bf Emoji Sentence Representations}: The main sentence representation we trained from emoji cluster model.

{\bf Emotional Word Vectors (EVEC)}: Average of emotional word vectors learned from hashtag corpus (300 dimension was selected through the performance in the dev set). See Chapter \ref{sec-word-rep} for more details.

{\bf Tweet-specific features}: We employ Tweet-specific features to capture information that two previous representations cannot. Inspired from the previous SemEval papers \cite{zhou2016ecnu,balikas2016twise}, we choose five features, (1) number of words in uppercase, (2) number of positive and negative emoticons, (3) Sum of emoji valence score \footnote{https://github.com/words/emoji-emotion}, (4) number of elongated words, and (5) number of exclamation and question marks. Note that we do not use any linguistic features or sentiment/emotion lexicons redundant to the two feature types above.

Experiments were implemented by using PyTorch \cite{paszke2017automatic} and Scikit-learn \cite{pedregosa2011scikit}. We used the given development set to tune the hyper-parameters and select models. For the final submission, we merged the train and development set together to retrain the model with the best hyper-parameter found (Table \ref{table:dataset}).

\section{Results and Discussion}

\subsection{Emoji Prediction}

As a result, the model achieved 29.8\% top-1 accuracy and 61.0\% top-3 accuracy on the emoji cluster prediction task. Since the objective is not to predict the cluster label but to find a good sentence representation, we visualized the test set samples to discover that samples with similar semantics and emotions were grouped together (Table \ref{table:emoji-eval}). 

\begin{table}[t]
\footnotesize
\begin{center}
\begin{tabular}{|l|}
\hline \bf One thing i dislike is laggers man \\ 
\hline
I hate inconsistency \\
The paper is irritating me \\
As of right now i hate dre \\
This is so unnecessary \\
my mom always ruining my great nights \\
\hline \bf im sick of crying im tired of trying \\ 
\hline
why body pain why \\
uuugh i really have nothing to do right now \\
i dont wanna go back to mex \\
i dont want to go to woooork \\
arghh back to reality now and im missing kl already \\
\hline \bf looking forward to holiday \\ 
\hline 
well today am on lake garda enjoying the life \\
perfect time to read book \\
im feeling great enjoying my holiday\\
coffee for every meal \#coffeelover \\
another beautiful day here at tynemouth \#bliss \\
\hline
\end{tabular}
\end{center}
\caption{\label{table:emoji-eval} Test samples from the emoji corpus and their top-5 nearest sentences according to the learned representations. Note that sentences with similar semantics and emotions are grouped together.}
\end{table}

\subsection{SemEval-2018 Results}

\begin{table*}[t]
\footnotesize
\begin{center}
\begin{tabular}{|l|l|cccc|c|}
\hline \multirow{3}{*}{\bf{Features}} & \multirow{3}{*}{\bf{Regression Method}} & \multicolumn{5}{c|}{Pearson correlation (all instances)}  \\  \cline{3-7}
& & \multicolumn{4}{c|}{\bf{1 (EI-reg)}} & \bf{3 (V-reg)} \\ \cline{3-7}
 &  & Anger & Fear & Joy & Sadness & Valence  \\ 
\hline
Word Unigram (baseline) & SVR &  .526 & .525 & .575 & .453 & .585 \\
Emoji Cluster & SVR & .733 & .632 & .679 & .693 & .811 \\
Emoji Cluster & KernelRidge & .735 & .638 & .675 & .692 & .809 \\
Emoji Cluster + EVEC & SVR & .739 & .678 & .701 & .706 & .815 \\
Emoji Cluster + EVEC & KernelRidge & .741 & .694 & .709 & .709 & .822 \\
Emoji Cluster + EVEC + features & SVR & \bf{.757} & .684 & \bf{.720} & \bf{.725} & \bf{.844} \\
Emoji Cluster + EVEC + features & KernelRidge & \bf{.757} & \bf{.698} & .693 & .721 & .840 \\
\hline
\end{tabular}
\end{center}
\caption{\label{table:reg-result} Test set results on Subtask 1/3. For Subtask 1, separate regression models were trained for each emotion category. Combining all three features to train shows the best performance }
\end{table*}

\subsubsection{Regression: Subtask 1/3}
\label{sec:reg-result}

\begin{table*}[h]
\footnotesize
\begin{center}
\begin{tabular}{|l|l|cccc|c|}
\hline \multirow{3}{*}{\bf{Features}} & \multirow{3}{*}{\bf{Regression Method}} & \multicolumn{5}{c|}{Pearson correlation (all instances)}  \\  \cline{3-7}
& & \multicolumn{4}{c|}{\bf{1 (EI-reg)}} & \bf{3(V-reg)} \\ \cline{3-7}
 &  & Anger & Fear & Joy & Sadness & Valence  \\ 
\hline Emoji Cluster & SVR & .733 & .632 & .679 & .693 & .811 \\
Emoji Cluster & KernelRidge & .735 & .638 & .675 & .692 & .809 \\
DeepMoji & SVR & .772 & .675 & .736 & .664 & .798 \\
DeepMoji & KernelRidge & .778 & .672 & .737 & .698 & .798 \\
Emoji Cluster + EVEC & SVR & .739 & .678 & .701 & .706 & .815 \\
Emoji Cluster + EVEC & KernelRidge & .741 & .694 & .709 & .709 & .822 \\
DeepMoji + EVEC & SVR & .781 & .694 & \underline{.749} & .708 & .810 \\
DeepMoji + EVEC & KernelRidge & .779 & .702 & \underline{.754} & .710 & .813 \\
DeepMoji + feat. & SVR & \underline{.785} & .680 & .739 & .714 & .824 \\
DeepMoji + feat. & KernelRidge & \underline{.781} & .670 & .691 & .711 & \underline{.829} \\
Emoji Cluster + EVEC + features & SVR & \underline{.757} & \underline{.684} & \underline{.720} & \underline{.725} & \bf{\underline{.844}} \\
Emoji Cluster + EVEC + features & KernelRidge & \underline{.757} & \underline{.698} & \underline{.693} & \underline{.721} & \underline{.840} \\
DeepMoji + EVEC + features & SVR & \bf{.792} & .709 & \bf{.763} & .732 & .837 \\
DeepMoji + EVEC + features & KernelRidge & .790 & \underline{\bf{.716}} & .734 & \underline{\bf{.739}} & .826 \\
\hline \multicolumn{2}{|l|}{\bf{Best Ensemble}} & \underline{\bf{.811(2)}} & \underline{\bf{.728(7)}} & \underline{\bf{.773(2)}} & \underline{\bf{.753(5)}} & \underline{\bf{.860(3)}} \\
\hline
\end{tabular}
\end{center}
\caption{\label{table:reg-result-all} Test set results on Subtask 1/3. The number next to the best ensemble(bold and underlined) indicates our ranking in the competition. Underlined ones show the models that were selected for ensemble according to the dev set. Ensembling models show a big boost to the performance.}
\end{table*}

Table \ref{table:reg-result} shows the test set results on regression tasks, Subtask 1/3, only using our emoji representation, Emoji Cluster. Our proposed representations significantly outperform the baseline. EVEC overall did help enhance the performance of the regression model for all emotion categories. This shows that emotional word vectors can serve as additional word-level information which are helpful for solving this task. Tweet-specific features boosted the performance, notably for sentiment, since features like capital letters, emojis, elongated words, and the number of exclamation marks, could help to figure out the subtle difference of the emotion intensities.

Table \ref{table:reg-result-all} also show results of all the representations we used for the competition. \textit{emoji cluster} worked better on sadness and sentiment, whereas \textit{DeepMoji} outperformed in anger, fear, and joy. We presumed such difference was due to the different emoji types of the two datasets used to train each model. \textit{Emoji cluster} only used 11 classes of emojis that were clustered together, but \textit{DeepMoji} used 64 emoji classes. It may be possible clustering of emoji classes made it easy for regression models to predict the intensities in certain emotion categories, whereas some emotion categories needed more detailed representations. In the end, we ensembled models using both representations to achieve a higher performance in the competition. Which model to ensemble was decided by the performance on the dev set.

\subsubsection{Ordinal Classification: Subtask 2/4}
\label{sec:ordinal-result}

Since we used the ensemble of the best regression models from above, we significantly outperformed the baseline. Finding the mapping between regression and ordinal labels turned out to be very effective method to do ordinal classification using regression models. 

Due to the fact that the datasets of regression tasks (EI-reg \& V-reg) and ordinal classification tasks (EI-oc \& V-oc) have the same sample sentences, we assume that regression labels are more informative than the ordinals, since they tell us the rank among the samples within the same ordinal class. Therefore, we first train a regression model and then use it to predict ordinals, rather than training a separate classifier. We later prove that this trick yields a better result in ordinal classification.

For regression, since our features are extracted from deep learning models, we find Support Vector Regression (SVR) and Kernel-Ridge Regression methods, which are effective for nonlinear features, perform better than linear methods. We tune the hyper-parameters with the given development (dev) set and later merge both train and dev set to train the final model with the best hyper-parameter found. Also, we try ensembles by averaging the final regression predictions of different methods or feature combinations to boost performance. The best groups of models are selected by the development set results of many combinations. 

\begin{figure}[h]
  \centering
  \includegraphics[width=6cm]{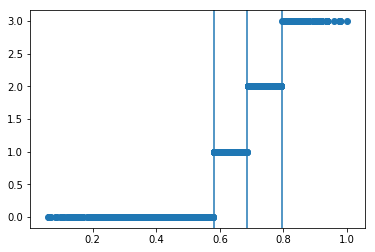}
  \caption{Distribution of regression labels (x-axis) and ordinal labels (y-axis) on the training dataset of Task 1a \& 2a for emotion category, fear. Class 0 for fear is distributed in [0,0.6]}
  \label{fig:fear_dist}
\end{figure}

\begin{figure}[h]
\centering
\includegraphics[width=6cm]{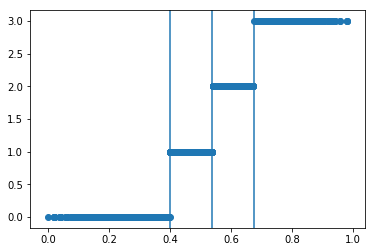}
\caption{Distribution of regression labels (x-axis) and ordinal labels (y-axis) on the training dataset of Task 1 \& 2. class 0 for joy is distributed in [0, 0.4]}
\label{fig:joy_dist}
\end{figure}

Another important finding is that the mapping between the regression labels and ordinal labels are very different among emotion categories. For example in Figure \ref{fig:fear_dist} and \ref{fig:joy_dist}, Class 0 for fear is distributed in [0,0.6], whereas class 0 for joy is distributed in [0, 0.4]. Therefore, we try to find the mapping from the regression values (continuous) to ordinal values (discrete) from the training dataset. We experiment with three different mapping:

\begin{enumerate}
\item \textit{naive mapping}: divides [0,1] into same size segments according to the number of ordinals
\item \textit{scope mapping}: finds the boundary of each segment in the training dataset (vertical lines on Figure \ref{fig:joy_dist} and \ref{fig:fear_dist})
\item \textit{polynomial mapping}: fits a polynomial regression function from the training data and finds the closest ordinal label. 
\end{enumerate}

\subsubsection{Results on Subtask 2/4 with ensembles}

\begin{table}[h]
\footnotesize
\begin{center}
\begin{tabular}{|l|r|ccc|}
\hline \multicolumn{2}{|c|}{} & \multicolumn{3}{c|}{Pearson (all instances)} \\ \cline{3-5}
\multicolumn{2}{|c|}{\bf{Task}} & Naive & Scope & Poly \\
\hline \multirow{5}{*}{2a (EI-oc)} & Anger & .654 & .664 & \bf{.704(2)} \\
 & Fear & .498 & .562 & \bf{.570(*)} \\
 & Joy & .632 & \bf{.720(1)} & .712 \\
 & Sadness & .645 & \bf{.697(*)} & .692 \\
\hline 4a (V-oc) & Valence & .813 & .816 & \bf{.833(2)} \\
\hline
\end{tabular}
\end{center}
\caption{\label{table:oc-result} Test set results on Subtask 2a \& 4a. The predictions of the best regression models are mapped into ordinal predictions. The number next to the best result(bold \& underlined) indicates our ranking of the competition. (*) indicates better results that we acquired after our final submission}
\end{table}

We used our best regression model to also predict ordinal labels. Since each emotion category has a different distribution of regression labels and ordinal labels, we experimented three different mappings, \textit{naive mapping}, \textit{scope mapping}, and \textit{polynomial mapping}. Using the training set, we found the ideal mapping function to match the regression predictions and the ordinal predictions. 

\begin{figure}[t]
\centering
\includegraphics[width=7cm]{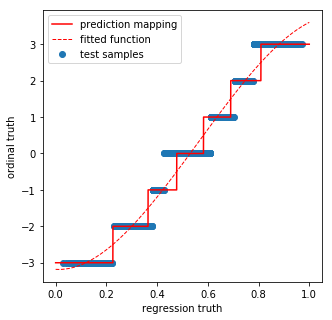}
\caption{Plot of test labels and the mapping function derived from the training set. A polynomial  function is fitted to map the regression predictions into ordinal predictions}
\label{fig:v_test_map}
\end{figure}

Test set results (Table \ref{table:oc-result}) on ordinal classification show that our mapping methods are indeed much more effective. For anger, fear, and sentiment categories, polynomial mapping performed the best, whereas scope mapping outperformed for joy and sadness categories. With our method, we achieved higher ranks in ordinal classification tasks (2a \& 4a), placed both in 2nd. Figure \ref{fig:v_test_map} shows how a cubic function is fitted to find the mapping between regression labels and ordinal labels. 

Additionally, we report some results better than the final submission. The change is due to a new model selection strategy. For the final submission, we searched for the optimal pair of regression model \& mapping method by looking at the ordinal classification results on the development set. However, it turned out that always using the best ensemble prediction and then searching for the optimal mapping method with respect to the development set was better.

\subsubsection{Multi-label Classification: Subtask 5}
\label{sec:multi-result}

This task of multi-label classification is different from previous tasks in that the model needs to predict the binary label for each of the 11 classes given a tweet. The task is difficult in terms of three aspects. Firstly, some of the classes have opposite emotions (such as optimism and pessimism) but may have been labeled both as true. Secondly, it is not trivial to distinguish similar emotions such as joy, love, and optimism, which will include a lot of noise in the labels and make it hard to perform classification during training. Lastly, most of the tweets are labeled with no more than 3 categories out of 11 classes, which make the labels very sparse and imbalanced (Table \ref{table:label_dist} ).

We propose to train two models to tackle this problem: regularized linear regression and logistic regression classifier chain \cite{read2009classifier}. Both models aim to exploit labels' correlation to perform multi-label classification. 

{\noindent \bf Regularized linear regression model}

We formulate the multi-label classification problem as a linear regression with label distance as the regularization term. We denote the features for i-th tweet as $x_i \in R^N $ where N is the number of features and the number of categories as C. Our prediction is $y^{'}_i = W * x_i$ where $W \in R^{M*C}$ is the weight of the linear regression model. We take the following formula as loss function to minimize. The loss consists of two parts. First part aims to minimize the mean square loss between our prediction $y^{'}_i$ and ground truth label $y_i$. The second part is the regularization term to capture relationship among different emotion labels. To model the correlations among emotions, we implicitly treat each emotion category as a vertice in an undirected graph g and use Laplacian matrix of g for regularization \cite{grone1990laplacian, shahid2016pca} . 

\begin{align*}
loss &= \frac{1}{M}\sum_i^M (y' - y)^2 + \lambda y^{'T}_i L y^{'}_i \\
L &= D - A 
\end{align*}

where M is the number of samples,  $L \in R^{C*C}$ is the Laplacian matrix, $A \in R^{C*C}$ is the Euclidean matrix, $D \in R^{C*C}$ is the Degree matrix. To derive L, we first compute the co-occurrence matrix $O \in R^{C*C}$ among the emotion labels and take each row/column $O_i \in R^C$ as the representation of each emotion. Then we compute the distance matrix A  by taking the Euclidean distance of different labels. That is  $A_{ij} = (O_i - O_j)^2 $. Here, A can be regarded as the adjacency matrix of the graph g. Afterwards, we calculate the degree matrix D by summing up each row/column and making it a diagonal matrix.  

{\noindent \bf Logistic regression classifier chain}
Classifier chain is another method to capture the correlation of emotion labels. It treats the multi-label problem as a sequence of binary classification problem while taking the prediction of the previous classifier as extra input. For example, when training the i-th emotion category, we take both the features of input tweet and also the 1st, 2nd, $\cdots$, (i-1)-th prediction as the input of our logistic regression classifier to predict the i-th emotion label of input tweet. We further ensemble 10 logistic regression chains by shuffling the sequence of 11 emotion labels to achieve better generalization ability.

\begin{table}[h]
\footnotesize
\begin{center}
\begin{tabular}{|l|cc|}
\hline \bf{Features} & \bf{Classifier Chain} & \bf{Regularized LR}  \\
\hline Word Unigram Baseline  & \multicolumn{2}{c|}{.442} \\ 
Emoji Cluster & .528 & .545  \\
Emoji Cluster + EVEC & .545 & .558  \\
Emoji Cluster + EVEC + features & \bf{.546}  & \bf{.558} \\
\hline
\end{tabular}
\end{center}
\caption{\label{table:clf-result} Test set results on Subtask 5. The competition metric is Jaccard index. Just like Subtask 1 and 2, combining all three features shows the best performance.}
\end{table}

Subtask 5 is to predict the emotion category given a tweet. Since each tweet can have multiple labels, among 11 categories, this problem can be set as a multi-label classification. We propose using two methods, classifier chain and regularized linear regression (See details at Table \ref{table:clf-result} shows the test set results on Subtask 5. Our proposed representations and methodologies again significantly outperform the baseline. Similar to the previous tasks, combining emoji sentence representation and EVEC improve the performance of both models. 

\subsubsection{Final Ranking of SemEval2018}

We participated the official SemEval2018 competition based on the two representations we explored in this thesis. By achieving a high ranking among 35 teams in the competition (Top3 for all five English subtasks), we proved the effectiveness of our representation learning methodologies. 

One thing to note is that our system's rank in the fear category (7th) is relatively lower than other emotion categories (See Table \ref{table:reg-result-all}). We found out from the previous literature \cite{wood2016ruder} that fear emojis were the most ambiguous, having the least correlation with human-annotated emotion labels among the six emotion categories. On the other hand, joy emojis were the most highly correlated. This may explain our best performance in the joy category and worst performance in the fear category. Future models using emojis as a dataset may need to take this shortcoming into account.

\begin{table}[h]
\footnotesize
\begin{center}
\begin{tabular}{|c|c|c|}
\hline \bf{Subtask} & \bf{System} & \bf{Score(rank)} \\
\hline
\multirow{4}{*}{\bf{1 EI-reg}}  & SeerNet & .799(1) \\
& NTUA-SLP & .776(2) \\
& \underline{\bf{PlusEmo2Vec}} & .766(3) \\
& psyML & .765(4) \\
\hline
\multirow{3}{*}{\bf{2} EI-oc}  & SeerNet & .695(1) \\
& \underline{\bf{PlusEmo2Vec}} & .659(2) \\
& psyML & .653(3) \\
\hline
\multirow{4}{*}{\bf{3 V-reg}}  & SeerNet & .873(1) \\
& TCS Research & .861(2) \\
& \underline{\bf{PlusEmo2Vec}} & .860(3) \\
& NTUA-SLP & .851(4) \\
\hline
\multirow{3}{*}{\bf{4} V-oc}  & SeerNet & .836(1) \\
& \underline{\bf{PlusEmo2Vec}} & .833(2) \\
& Amobee & .813(3) \\
\hline
\multirow{4}{*}{\bf{5 E-c}}  & NTUA-SLP & .588(1) \\
& TCS Research & .582(2) \\
& \underline{\bf{PlusEmo2Vec}} & .576(3) \\
& psyML & .574(4) \\
\hline

\end{tabular}
\end{center}
\caption{\label{table:ranking} Official final scoreboard on all 5 subtasks that we participated. Scores for Subtask 1-4 are macro-average of the Pearson scores of 4 emotion categories and 5 is Jaccard index. About 35 participants are in each task.}
\end{table}

\section{Conclusion}

In this chapter, we explored learning a good sentence-level representations of emotions in sentences by using a deep neural network, bidirectional LSTM, and a huge amount of corpus distantly labeled by emojis. As \cite{felbo2017using} suggest, modeling representations of emotions in sentences with emoji is a very effective method, especially for working with affect-related tasks. We replicated their work by using a much smaller dataset (1.5\% of 1.2B) and less number of model parameters. Our model can achieve top-1 accuracy of 30\% and top-3 accuracy of 61\% to classify which emoji cluster the sentence is labeled to. Although this accuracy is not perfect, the qualitative analysis of the learned representations shows that the model is able to capture the similarity of emotions among sentences.

To further prove the significance of our representations, We evaluated our approach on sentiment/emotion intensity regression and multi-label classification task from SemEval-2018 Task 1 competition and proved that our emoji sentence representations with SVR or Kernel Ridge linear regression outperform the baseline by a big margin. Moreover, we showed that combining the word-level representations, EVEC, and emoji sentence representations is very effective. 

Moreover, we competed in the main competition with the representation learning approaches proposed so far and ranked among the top 3 in every five subtask of the competition we participated. Such result demonstrates that the performance of our approach is very capable, even among many other people's approaches proposed in the research community. Therefore, most results of this chapter have been published in our paper \cite{park2018plusemo2vec}.

%% file: chapter/sec-abusive.tex
\chapter{Abusive Language Detection}\label{sec-abusive}

In this chapter, we shift our focus to automatic detection of abusive language. It is also known as toxic comments or hate speech. Fighting abusive language online is becoming more and more important in a world where online social media plays a significant role in shaping the minds of people \cite{perse}. Nevertheless, major social media companies like Twitter find it difficult to tackle this problem \cite{meyer}, as a huge number of posts produced every day cannot be moderated with only human efforts.

The major difficulties of tackling abusive language is that relevant datasets are often very noisy, with many spelling errors and implicit slangs, and very hard to interpret because these words are often subjective to individuals \cite{ross} and lack clear standards for annotation. These attributes make it very difficult for non-experts to annotate without having a certain amount of knowledge and cause different corpora to have different labels, such as abusive, offensive, toxic, insult, etc. \cite{schmidt,waseemhovy} Another obstacle is that there are various domains that abusive language can appear. For example, abusive languages can appear in public forum posts and comments, social media posts and comments, on-line game chats, public speech, etc., with similar or distinct forms. Also, specific kinds of abusive language like sexism, racism, or personal threat, have yet been addressed much in previous works \cite{wassem}. 

Abusive language is relevant to our previous chapters of modeling emotions in that it is provoked from negative emotions such as anger and hatred and the format of the problem is text classification. We focus on using deep learning models that can automatically capture which representations are important (also known as feature learning \cite{bengio2013representation}) and finally classify whether the given text is abusive language or not. There are two main parts for this section.

We show that both word-level and character-level features are important in abusive language detection. We first propose convolutional neural network (CNN) models to learn the representations of abusive language from a large human-annotated corpus and predict whether a text is abusive or not, exploring both word-level and character-level representations to handle noisy words. Also, we propose a two-step approach of performing classification on abusive language and then classifying into specific types and compares it with one-step approach of doing one multi-class classification for detecting sexist and racist languages.

Secondly, we argue that existing machine learning approaches can generate unfairly biased representations. We address the issue of model bias and run experiments specific to gender identities (gender bias) to identify biases existing in the representations learned from abusive language datasets. We focus on the fact that the representations of abusive language learned in only supervised learning setting may not be able to generalize well enough for practical use since they tend to overfit to certain words that are neutral but occur frequently in the training samples. Without achieving certain level of generalization ability, abusive language detection models may not be suitable for real-life situations. We introduce ways to measure those model bias and apply three different methods - debiasing word embedding, data augmentation, and finetuning from bigger corpus - to mitigate them.

\section{Background}

\subsection{Definition of Abusive Language}
Many have addressed the difficulty of the definition of abusive language while annotating the data, because they are often subjective to individuals \cite{ross2017measuring} and lack of context \cite{waseemhovy,schmidt}. This makes it harder for non-experts to annotate without having a certain amount of domain knowledge \cite{wassem}. Recent work \cite{founta2018large} studied various formats of abusive language appearing on Twitter and found that the terms ``abusive'' and ``hate speech'' were the least redundant and representative labels while systematically annotating tweets for abusive language on Twitter. They summarized the definition of these terms employed in previous literatures.

\noindent {\bf Abusive Language}: ``Any strongly impolite, rude or hurtful language using profanity, that can show a debasement of someone or something, or show intense emotion.'' \cite{nobata2016abusive,papegnies2017detection}

\noindent {\bf Hate Speech}: ``Language used to express hatred towards a targeted individual or group, or is intended to be derogatory, to humiliate, or to insult the members of the group, on the basis of attributes such as race, religion, ethnic origin, sexual orientation, disability, or gender.'' \cite{badjatiya2017deep,davidson2017hateoffensive, djuric2015hate,schmidt,warner2012detecting}

\subsection{Related works}

So far, many efforts were put into defining and constructing abusive language datasets from different sources and labeling them through crowd-sourcing or user moderation \cite{davidson2017hateoffensive, founta2018large,wassem,waseemhovy, wulczyn2017ex}. Many deep learning approaches have been explored to train a classifier with those datasets to develop an automatic abusive language detection system \cite{badjatiya2017deep,pavlopoulos2017deeper, wassem, wulczyn2017ex}. 

However, these works do not explicitly address any model bias in their models. Addressing biases in NLP models/systems have recently started to gain more interest in the research community, not only because fairness in AI is important but also because bias correction can improve the robustness of the models. \cite{bolukbasi2016man} is one of the first works to point out the gender stereotypes inside word2vec \cite{mikolov2013distributed} and propose an algorithm to correct them. \cite{caliskan2017semantics} also propose a method called Word Embedding Association Test (WEAT) to measure model bias inside word embeddings and finds that many of those pretrained embeddings contain problematic bias toward gender or race. \cite{dixon2017measuring} is one of the first works that point out existing ``unintended'' bias in abusive language detection models. \cite{kiritchenko2018examining} compare 219 sentiment analysis systems participating in SemEval competition with their proposed dataset, which can be used for evaluating racial and gender bias of those systems. 

\section{Two-step Classification with CNN}

In this section, we aim to experiment a two-step approach of detecting abusive language first and then classifying into specific types and compare with a one-step approach of doing one multi-class classification on sexist and racist language. 

We use three kinds of CNN models that use both character-level and
word-level inputs to perform classification on different dataset segmentations. We measure the performance and ability of each model to capture characteristics of abusive language. 

\subsection{Methodology}

We propose to implement three CNN-based models to classify sexist and racist abusive language: CharCNN, WordCNN, and HybridCNN. The major difference among these models is whether the input features are characters, words, or both. The key components are the convolutional layers that each computes a one-dimensional convolution over the previous input with multiple filter sizes and large feature map sizes. Having different filter sizes is the same as looking at a sentence with different windows simultaneously. Max-pooling is performed after the convolution to capture the feature that is most significant to the output.

\subsubsection{CharCNN}

CharCNN is a modification of the character-level convolutional network in \cite{zhang2015character}. Each character in the input sentence is first transformed into a one-hot encoding of 70 characters, including 26 English letters, 10 digits, 33 other characters, and a newline character (punctuations and special characters). All other non-standard characters are removed. 

\cite{zhang2015character} uses 7 layers of convolutions and max-pooling layers, 2 fully-connected layers, and 1 softmax layer, but we also designed a shallow version with 2 convolutions and max-pooling layers, 1 fully-connected layers, and 1 softmax layers with dropout, due to the relatively small size of our dataset to prevent overfitting.

\subsubsection{WordCNN}

WordCNN is a CNN-static version proposed by \cite{Kim2014}. The input sentence is first segmented into words and converted into a 300-dimensional embedding word2vec trained on 100 billion words from Google News \cite{mikolov2013distributed}. Incorporating pretrained vectors is a widely-used method to improve performance, especially when using a relatively small dataset. We set the embedding to be non-trainable since our dataset is small.

\subsubsection{HybridCNN}

\begin{figure}[h]
  \centering
  \includegraphics[width=7cm]{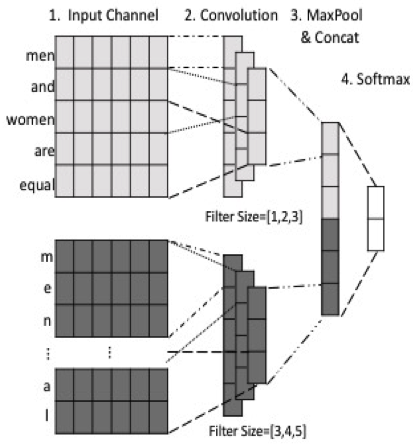}
  \caption{Architecture of HybridCNN}
  \label{fig:hybrid_cnn}
\end{figure}

We design HybridCNN, a variation of WordCNN, since WordCNN has the limitation of only taking word features as input. Abusive language often contains either purposely or mistakenly mis-spelled words and made-up vocabularies such as \#feminazi. Therefore, since CharCNN and WordCNN do not use character and word inputs at the same time, we design the HybridCNN to experiment whether the model can capture features from both levels of inputs.

HybridCNN has two input channels (Figure \ref{fig:hybrid_cnn}). Each channel is fed into convolutional layers with three filter windows of different size. The output of the convolution are concatenated into one vector after 1-max-pooling. The vector is then fed into the final softmax layer to perform classification.

\subsection{Experimental Setup}

\subsubsection{Hyperparameters for CNN models}
\label{cnn-hyperparm}

For hyperparameter tuning, we evaluated on the validation set. These are the hyperparmeters used for evaluation.
\begin{enumerate}
\item CharCNN: Shallow model with 1024 feature units for convolution layers with filter size 4, max-pooling size 3, and L2 regularization constant 1 and 2048 units for the fully-connected layer
\item  WordCNN: Convolution layers with 3 filters with the size of [1,2,3] and feature map size 50, max-pooling, and L2 regularization constant 1
\item  HybridCNN: For the character input channel, convolution layers with 3 filters with size of [3,4,5] and for word input channel, 3 filters with size of [1,2,3]. Both channels had feature map size of 50, max-pooling, and L2 regularization constant 1.  
\end{enumerate}

\subsubsection{Dataset}

\begin{table}[t]
\footnotesize
\begin{center}
\begin{tabular}{|c|ccc|cc|cc|}
\hline
Dataset & \multicolumn{3}{c}{\bf{One-step}} & \multicolumn{2}{|c}{\bf{Two-step-1}} & \multicolumn{2}{|c|}{\bf{Two-step-2}} \\
\hline 
Label & None & Racism & Sexism & None & Abusive & Sexism & Racism \\
\hline
\# & 12,427 & 2,059	& 3,864 & 12,427 & 5,923 & 2,059 & 3,864 \\
\hline
\end{tabular}
\end{center}
\caption{\label{table:alw_dataset} Dataset Segmentation}
\end{table}

We used the two English Twitter Datasets \cite{wassem,waseemhovy} published as unshared tasks for the 1st Workshop on Abusive Language Online(ALW1). It contains tweets with sexist and racist comments. \cite{waseemhovy} created a list of criteria based on a critical race theory and let an expert annotate the corpus. First, we concatenated the two datasets into one and then divided that into three datasets for one-step and two-step classification (Table \ref{table:alw_dataset} ). One-step dataset is a segmentation for multi-class classification. For two-step classification, we merged the sexism and racism labels into one abusive label. Finally, we created another dataset with abusive languages to experiment a second classifier to distinguish `sexism' and `racism', given that the instance is classified as `abusive'. 

\subsubsection{Training}

We performed two classification experiments: 

\begin{enumerate}
\item Detecting `none', `sexist', and `racist' language \textit{(one-step)}
\item Detecting `abusive language', then further classifying into `sexist' or `racist' \textit{(two-step)}
\end{enumerate}	

The purpose of these experiments was to see whether dividing the problem space into two steps makes the detection more effective.

We trained the models using mini-batch stochastic gradient descent with Adam \cite{Kingma2015}, the mini-batch with a size of 32 had been sampled with equal distribution for all labels. The training continued until the evaluation set loss did not decrease any longer. All the results are average results of 10-fold cross validation. Hyperparameters are found using the validation set.

\subsubsection{Evaluation}

As evaluation metric, we used F1 scores with precision and recall score and weighted averaged the scores to consider the imbalance of the labels. For this reason, total average F1 might not be-tween average precision and recall. 

As baseline, we used the character n-gram logistic regression classifier (indicated as LR on Table 2-4) \cite{waseemhovy}, Support Vector Machines (SVM) classifier, and FastText \cite{joulin2016bag} that uses average bag-of-words representations to classify sentences. It was the second best single model on the same dataset after CNN \cite{badjatiya2017deep}.

\subsection{Result and Discussion}

\begin{table}[h]
\footnotesize
\begin{center}
\begin{tabular}{|r|ccc|ccc|ccc||ccc|}
\hline
Label & \multicolumn{3}{c|}{\bf{None}} & \multicolumn{3}{c|}{\bf{Racism}} & \multicolumn{3}{c||}{\bf{Sexism}} & \multicolumn{3}{|c|}{\bf{Total}} \\
\hline Method & Prec. & Rec. & F1 & Prec. & Rec. & F1 & Prec. & Rec. & F1 & Prec. & Rec. & F1 \\
\hline 
LR & .824  & .945 & .881 & .810 & .598 & .687 & .835 & .556 & .668 & .825 & .824 & .814 \\
SVM & .802 & \underline{\bf{.956}} & .872 & \underline{\bf{.815}} & .531 & .643 & \underline{\bf{.851}} & .483 & .616 & .814 & .808 & .793 \\
FastText & .828& .922& .882& .759& .630& .685& .777& .557& .648& .810& .812& .804 \\
CharCNN&.861&.867&.864&.693&.746&.718&.713&.666&.688&.801&.811&.811 \\
WordCNN&.870&.868&.868&.704&.762&.731&.712&\bf{.686}&.694&.818&.816&.816 \\
HybridCNN& \bf{.872}&.882&.877&.713&\underline{\bf{.766}}&\underline{\bf{.736}}&.743&.679&\underline{\bf{.709}}&\bf{.827}&\bf{.827}&\bf{.827} \\
\hline
LR(2)&.841 &.933&\underline{\bf{.895}}&.800&.664&.731&.809&.590&.683&\underline{\bf{.828}}&\underline{\bf{.831}}&\bf{.824} \\
SVM(2)&.816&\bf{.945}&.876&\bf{.811}&.605&.689&\bf{.823}&.511&.630&.816&.815&.803 \\
HybridCNN(2) &.877&.864&.869&.690&\bf{.759}&.721&.705&.701&\bf{.699}&.807&.809&.807 \\
HybridCNN+LR(2)&\underline{\bf{.880}}&.859&.869&.722&.751&\bf{.735}&.683&\underline{\bf{.717}}&\bf{.699}&.821&.817&.818 \\
\hline
\end{tabular}
\end{center}
\caption{\label{table:total-results}Experiment Results: upper group is the one-step methods that perform multi-class classification and lower group with (2) indicates two-step that combines two binary classifiers. HybridCNN, our newly created model, with one-step method shows the best performance in `racism' and `sexism' labels, but surprisingly logistic regression (LR) with two-step method shows the best performance for total F1 score.}
\end{table}

\subsubsection{One-step Classification}

The results of the one-step multi-class classification are shown in the top part of Table \ref{table:total-results}. Our newly proposed HybridCNN performs the best, giving an improvement over the result from WordCNN. We expected the additional character input channel improves the performance. We assumed that the reason CharCNN performing worse than WordCNN is that the dataset is too small for the character-based model to capture word-level features by itself. Baseline methods tend to have high averaged F1 but low scores on racism and sexism labels due to low recall scores. 

\subsubsection{Two-step Classification}

Two-step approach that combines two binary classifiers shows comparable results with one-step approach. The results of combining are shown in the bottom part of Table \ref{table:total-results}. 

Combining two logistic regression classifiers in the two-step approach performs about as well as one-step HybridCNN and outperform one-step logistic regression classifier by more than 0.1 F1 points (Table \ref{table:total-results}). This is surprising since logistic regression takes less features than the HybridCNN. Furthermore, using HybridCNN on the first step to detect abusive language and logistic regression on the second step to classify racism and sexism worked better than just using HybridCNN.

\noindent\begin{minipage}{.45\linewidth}
   \centering
\begin{tabular}{|r|ccc|}
\hline 
Model & Prec. & Rec. & F1 \\
\hline
LR & .816 & .640 & .711 \\
SVM & \bf{.839} & .560 & .668 \\
FastText & .765 & .616 & .683 \\
CharCNN & .743 & .674 & .707 \\
WordCNN & .731 & .722 & .726 \\
HybridCNN & .719 & \bf{.754} & \bf{.734} \\
\hline
\end{tabular}
\captionof{table}{Results on Abusive Language Classification (first-step). This is to classify whether the tweet is `abusive' (either `sexist' or `racist') or not. HybridCNN shows the best performance in terms of F1 score.}
\label{table:two-step-1}

\end{minipage}\hfill
\begin{minipage}{.45\linewidth}

\begin{tabular}{|r|ccc|}
\hline 
Model & Prec. & Rec. & F1 \\
\hline
LR & \bf{.954} & \bf{.953} & \bf{.952} \\
SVM & \bf{.954} & \bf{.953} & \bf{.952} \\
FastText & .937 & .937 & .937 \\
CharCNN & .941 & .941 & .941 \\
WordCNN & .952 & .952 & .\bf{952} \\
HybridCNN & .951 & .950 & .950 \\
\hline
\end{tabular}
\captionof{table}{Results on Sexist/Racist Classification (second-step). Even simple methods like logistic regression (LR) and support vector machines (SVM) show comparable or better results than CNN models.}
\label{table:two-step-2}
\end{minipage}

Table \ref{table:two-step-1} shows the results of abusive language classification. HybridCNN also performs best for abusive language detection, followed by WordCNN and logistic regression. 

Table \ref{table:two-step-2} shows the results of classifying into sexism and racism given that it is abusive. The second classifier has significant performance in predicting a specific type (in this case, sexism and racism) of an abusive language. We can deduce that sexist and racist comments have obvious discriminating features that are easy for all classifiers to capture. 

Since the precision and recall scores of the abusive label is higher than those of racism and sexism in the one-step approach, the two-step approach can perform as well as the one-step approach. With many different machine learning classifiers including our proposed HybridCNN, which takes both character and word features as input, we showed the potential in the two-step approach compared to the one-step approach which is simply a multi-class classification. In this way, we can boost the performance of simpler models like logistic regression, which is faster and easier to train, and combine different types of classifiers like convolutional neural network and logistic regression together depending on each of its performance on different datasets. 

\section{Model Bias in Abusive Language Detection Models}

While working on the dataset in the previous section, we found out that the models are overfitting to certain words that are not the most important words for detecting abusive language. For example, sentences like ``You are a good woman'' are likely to be considered ``sexist'' because of the word ``woman.''. In this section, we directly address this problem, model bias in abusive language detection models. 

Although previous work tackled on using machine learning to automatically detect abusive language, recent works have raised concerns about the robustness of those systems. \cite{hosseini2017deceiving} have shown how to easily cause false predictions with adversarial examples in Google's API, and \cite{dixon2017measuring} show that classifiers can have unfair biases toward certain groups of people. 

This phenomenon, called \textit{false positive bias}, has been reported by \cite{dixon2017measuring}. They further defined this model bias as unintended, \textit{``a model contains unintended bias if it performs better for comments containing some particular identity terms than for comments containing others.''} 

Such model bias is important but often unmeasurable in the usual experiment settings since the validation/test sets we use for evaluation are already biased. For this reason, we first introduce methods to measure these model biases and then conduct experiments on how to reduce them.

\subsection{Measuring Gender Bias}
\subsubsection{Dataset}

We introduce two datasets for the upcoming experiments (Table \ref{table:dataset_stats}). We measure the model bias specific to gender identity terms in these two datasets.

\noindent {\bf{Sexist Tweets (\texttt{st})}}
\label{sec:waseem}
This dataset consists of tweets with sexist tweets collected from Twitter by searching for tweets that contain common terms pertaining to sexism such as ``feminazi.'' The tweets were then annotated by experts based on criteria founded in critical race theory. The original dataset also contained a relatively small number of ``racist'' label tweets, but we only retain ``sexist'' samples to focus on gender biases. \cite{wassem,waseemhovy}, the creators of the dataset, describe ``sexist'' and ``racist'' languages as specific subsets of abusive language. 

\noindent {\bf{Abusive Tweets (\texttt{abt})}}
\label{sec:founta}
Recently, \cite{founta2018large} has published a large scale crowdsourced abusive tweet dataset with 60K tweets. Their work incrementally and iteratively investigated methods such as boosted sampling and exploratory rounds, to effectively annotate tweets through crowdsourcing. Through such systematic processes, they identify the most relevant label set in identifying abusive behaviors in Twitter as $\{None, Spam, Abusive, Hateful\}$ resulting in 11\% as 'Abusive,' 7.5\% as 'Hateful', 22.5\% as 'Spam', and 59\% as 'None'. We transform this dataset for a binary classification problem by concatenating 'None'/'Spam' together, and 'Abusive'/'Hateful' together.

\begin{table}[t] 
	\centering
\begin{tabular}{@{}cccccc@{}}
	\toprule
Name & Size & Positives (\%) & $\mu$ & $\sigma$ & $max$ \\
	\midrule
Sexist Tweets (\texttt{st}) & 18K & 33\% & 15.6 & 6.8 & 39 \\ 
Abusive Tweets (\texttt{abt}) & 60K & 18.5\% & 17.9 & 4.6 & 65 \\ 
	\bottomrule
\end{tabular}
\caption{Dataset statistics. $\mu, \sigma, max$ are mean, std.dev, and maximum of sentence lengths} \label{table:dataset_stats}
\end{table}

\subsubsection{Methodology}

Gender bias cannot be measured when evaluated on the original dataset as the test sets will follow the same biased distribution, so normal evaluation set will not suffice. Therefore, we generate a separate \textit{unbiased test set} for each gender, male and female, using the identity term template method proposed in \cite{dixon2017measuring}. 

The intuition of this template method is that given a pair of sentences with only the identity terms different (ex. ``He is happy'' and ``She is happy''), the model should be able to generalize well and output same prediction for abusive language. This kind of evaluation has also been performed in \textit{SemEval 2018: Task 1 Affect In Tweets} \cite{kiritchenko2018examining} to measure the gender and race bias among the competing systems for sentiment/emotion analysis. They developed eleven templates with different identity terms and emotional state words to measure the racial and gender biases.

Using the released code\footnote{\url{https://github.com/conversationai/unintended-ml-bias-analysis}} of \cite{dixon2017measuring}, we generated 1,152 samples (576 pairs) by filling the templates with common gender identity pairs (ex. male/female, man/woman, etc.). We created templates (Table \ref{table:templates}) that contained both neutral and offensive nouns and adjectives (Table \ref{table:offensive-words}) inside the vocabulary to retain balance in neutral and abusive samples.

\begin{table}[t]
\footnotesize
\begin{center}
\begin{tabular}{|c|}
\hline \bf{Example Templates} \\ 
\hline
You are a (adjective) (identity term). \\
(verb) (identity term). \\
Being (identity term) is (adjective) \\
I am (identity term) \\
I hate (identity term) \\
\hline
\end{tabular}
\end{center}
\caption{\label{table:templates} Example of templates used to generated an unbiased test set.}
\end{table}

\begin{table}[t]
\footnotesize
\begin{center}
\begin{tabularx}{\textwidth}{|X|X|}
\hline Type & Example Words \\ 
\hline
Offensive & disgusting, filthy, nasty, rude, horrible, terrible, awful, worst, idiotic, stupid, dumb, ugly, shitty, fucked, pathetic, useless, shit, bad, crap, absurd, annoying, kill, murder, hate, fuck, assholes, rape, hypocrites \\
\hline
Non-offensive & help, love, respect, believe, congrats, hi, like, great, fun, nice, neat, happy, good, best, fantastic, wonderful, lovely, excellent, incredible, friendly, gracious, kind, better, best, interesting, healthy, liked, famous \\
\hline
\end{tabularx}
\end{center}
\caption{\label{table:offensive-words} Example of offensive and non-offensive words used in the generated test set.}
\end{table}

For the evaluation metric, we use 1) AUC scores on the original test set (Orig. AUC), 2) AUC scores on the unbiased generated test set (Gen. AUC), and 3) the false positive/negative equality differences proposed in \cite{dixon2017measuring} which aggregates the difference between the overall false positive/negative rate and gender-specific false positive/negative rate. False Positive Equality Difference (FPED) and False Negative Equality Difference (FNED) are defined as below, where $T = \{male,female\}$. 
\begin{gather*} 
FPED = \sum_{t\in T} | FPR - FPR_t| \\
FNED = \sum_{t\in T} | FNR - FNR_t| 
\end{gather*} 
Since the classifiers output probabilities, equal error rate thresholds are used for prediction decision. 

While the two AUC scores show the performances of the models in terms of accuracy, the equality difference scores show them in terms of fairness, which we believe is another dimension for evaluating the model's generalization ability. 

\subsubsection{Experimental Setup}
\label{sec: experiment}
We first measure gender biases in \texttt{st} and \texttt{abt} datasets. We explore three neural models used in previous works on abusive language classification: Convolutional Neural Network (CNN) \cite{park2017one}, Gated Recurrent Unit (GRU) \cite{cho2014learning}, and Bidirectional GRU with self-attention ($\alpha$-GRU) \cite{pavlopoulos2017deeper}, but with a simpler mechanism used in \cite{felbo2017using}.

We also compare different pre-trained embeddings, \texttt{word2vec} \cite{mikolov2013distributed} trained on Google News corpus, \texttt{FastText} \cite{joulin2016bag}) trained on Wikipedia corpus, and randomly initialized embeddings (\textit{random}) to analyze their effects on the biases. Experiments were run 10 times and averaged.

These are the hyperparameters for CNN, GRU, and $\alpha$-GRU:

\begin{enumerate}
\item CNN: Convolution layers with 3 filters with the size of [3,4,5], feature map size=100, Embedding Size=300, Max-pooling, Dropout=0.5
\item GRU: hidden dimension=512, Maximum Sequence Length=100, Embedding Size=300, Dropout=0.3
\item $\alpha$-GRU: hidden dimension=256 (bidirectional, so 512 in total), Maximum Sequence Length=100, Attention Size=512, Embedding Size=300, Dropout=0.3
\end{enumerate}

\subsubsection{Results \& Discussion}
\label{sec:measurement-results}
Tables \ref{table:waseem-result} and \ref{table:founta-result} show the bias measurement experiment results for \texttt{st} and \texttt{abt}, respectively. As expected, pre-trained embeddings improved task performance. The score on the unbiased generated test set (Gen. AUC) also improved since word embeddings can provide prior knowledge of words.

\begin{table}[t]
\begin{center}
\begin{tabular}{|l|l|c|c|c|c|}
\hline
\thead{\bf{Model}} & \thead{\bf{Embed.}} & \thead{\bf{Original AUC}}  & \thead{\bf{Gen. AUC}} & \thead{\bf{FNED}} & \thead{\bf{FPED}} \\
\hline \multirow{3}{*}{{CNN}}          & random & .881        & .572      & .261      & .249 \\
                                       & fasttext & .\bf{906} & .620      & .323      & .327 \\
                                       & word2vec & \bf{.906} & .635      & .305      & .263  \\
\hline \multirow{3}{*}{{GRU}}          & random & .854        & .536      & \bf{.132} & .\bf{136} \\
                                       & fasttext & .887      & \bf{.661} & .312      & .284 \\
                                       & word2vec & .887      & .633      & .301      & .254  \\
\hline \multirow{3}{*}{{$\alpha$-GRU}} & random   & .868      & .586      & .236      & .219 \\
                                       & fasttext & .891      & .639      & .324      & .365 \\
                                       & word2vec & .890      & .631      & .315      & .306  \\
\hline
\end{tabular}
\end{center}
\caption{\label{table:waseem-result} Results on \texttt{st}. False negative/positive equality differences are larger when pre-trained embedding is used and CNN or $\alpha$-RNN is trained}
\end{table}

\begin{table}[t]
\begin{center}
\begin{tabular}{|l|l|c|c|c|c|}
\hline
\thead{\bf{Model}} & \thead{\bf{Embed.}} & \thead{\bf{Original AUC}}  & \thead{\bf{Gen. AUC}} & \thead{\bf{FNED}} & \thead{\bf{FPED}} \\
\hline \multirow{3}{*}{{CNN}} & random & .926        & .893      & .013      & .045 \\
                              & fasttext & .955      & .995      & .004      & \bf{.001} \\
                              & word2vec & \bf{.956} & \bf{.999} & \bf{.002} & .021  \\
\hline \multirow{3}{*}{{GRU}} & random & .919        & .850      & .036      & .010 \\
                              & fasttext & .951      & .997      & .014      & .018 \\
                              & word2vec & .952      & .997      & .017      & .037  \\
\hline \multirow{3}{*}{{$\alpha$-GRU}} & random & .927      & .914      & .008      & .039 \\
                              & fasttext & \bf{.956} & .998      & .014      & .005 \\
                              & word2vec & .955      & \bf{.999} & .012      & .026  \\
\hline
\end{tabular}
\end{center}
\caption{\label{table:founta-result} Results on \texttt{abt}. The false negative/positive equality difference is significantly smaller than the \texttt{st}}
\end{table}

However, the equality difference scores tended to be larger when pre-trained embeddings were used, especially in the \texttt{st} dataset. This confirms the result of \cite{bolukbasi2016man,caliskan2017semantics}. In all experiments, direction of the gender bias was towards female identity words. We can infer that this is due to the more frequent appearances of female identities in ``sexist'' tweets and lack of negative samples, similar to the reports of \cite{dixon2017measuring}. This is problematic since not many NLP datasets are large enough to reflect the true data distribution, more prominent in tasks like abusive language where data collection and annotation are difficult.

On the other hand, \texttt{abt} dataset showed significantly better results on the two equality difference scores, of at most 0.04. Performance in the generated test set was better because the models successfully classify abusive samples regardless of the gender identity terms used. Hence, we can assume that \texttt{abt} dataset is less gender-biased than the \texttt{st} dataset, presumably due to its larger size, balance in classes, and systematic collection method. 

Interestingly, the architecture of the models also influenced the biases. Models that ``attend'' to certain words, such as CNN's max-pooling or $\alpha$-GRU's self-attention, tended to result in higher false positive equality difference scores in \texttt{st} dataset. These models show effectiveness in catching not only the discriminative features for classification, but also the ``unintended'' ones causing the model biases.

\subsection{Reducing Gender Bias}
We experiment various methods to reduce gender biases identified in Section \ref{sec:measurement-results}. 

\subsubsection{Methodology}

\noindent \textbf{Debiased Word Embeddings (DE)} - \cite{bolukbasi2016man} proposed an algorithm to correct word embeddings by removing gender stereotypical information. All the other experiments used pretrained word2vec to initialized the embedding layer but We substitute their published embeddings with the pretrained word2vec to verify their effectiveness in our task. 

\noindent \textbf{Gender Swap Data Augmentation(GS)} - We augment the training data by identifying male entities and swapping them with equivalent female entities and vice-versa. This simple method removes correlation between gender and classification decision and has proven to be effective for correcting gender biases in co-reference resolution task \cite{zhao2018gender}.

\noindent \textbf{Bias fine-tuning (FT)} - We propose a method to use transfer learning from a less biased corpus to reduce the bias. A model is initially trained with a larger, less-biased source corpus with a same or similar task, and fine-tuned with a target corpus with a larger bias. This method is inspired by the fact that model bias mainly rises from the imbalance of labels and the limited size of data samples. Training the model with a larger and less biased dataset may regularize and prevent the model from overfitting to the small, biased dataset.

\subsubsection{Experimental Setup}
\textit{Debiased word2vec} \cite{bolukbasi2016man} is compared with the original \textit{word2vec} \cite{mikolov2013distributed} for evaluation. For gender swapping data augmentation, we use pairs identified through crowd-sourcing by \cite{zhao2018gender}.

After identifying the degree of gender bias of each dataset, we select a source with less bias and a target with more bias. Vocabulary is extracted from training split of both sets. The model is first trained by the source dataset. We then remove final softmax layer and attach a new one initialized for training the target. The target is trained with a slower learning rate. Early stopping is decided by the valid set of the respective dataset.

Based on this criterion and results from Section \ref{sec:measurement-results}, we choose the \texttt{abt} dataset as source and \texttt{st} dataset as target for bias fine-tuning experiments. 

\subsubsection{Results \& Discussion}
\label{sec:analysis}

\begin{table}[h]
\begin{center}
\begin{tabular}{|c|ccc|c|c|c|c|}
\hline
\thead{\bf{Model}} & \thead{\bf{DE}} & \thead{\bf{GS}} & \thead{\bf{FT}} & \thead{\bf{Original AUC}}  & \thead{\bf{Gen. AUC}} & \thead{\bf{FNED}} & \thead{\bf{FPED}} \\
\hline \multirow{6}{*}{{CNN}} &  . & . & . &  \bf{\underline{.906}} & .635      & .305      & .263 \\
                               & O & . & .                 &  .902 & .627      & .333      & .337 \\
                               & . & O & .                 &  .898 & .676      & .164      & .104 \\
                               & O & O & .                 &  .895 & .647      & .157      & .096 \\                     
                               & . & . & O                 &  .896 & .650      & .302      & .240 \\
                               & . & O & O                 &  .889 & .671      & .163      & .122 \\
                               & O & O & O                 &  .884 & \underline{.703}      & \underline{.135} & \underline{.095} \\
\hline \multirow{6}{*}{{GRU}} &  . & . & .      &  \underline{.887} & .633      & .301      & .254 \\
                               & O & . & .                 &  .882 & .658      & .274      & .270 \\
                               & . & O & .                 &  .879 & .657      & .044      & .040 \\
                               & O & O & .                 &  .873 & .667      & \bf{\underline{.006}}      & \bf{\underline{.027}} \\                     
							   & . & . & O                 &  .874 & .761      & .241      & .181 \\
                               & . & O & O                 &  .862 & .768      & .141      & .095 \\                     
                               & O & O & O                 &  .854 & \underline{.854}      & .081 & .059 \\
\hline \multirow{6}{*}{{$\alpha$-GRU}}&  . & . & .& \underline{.890} & .631      & .315       & .306 \\
                               & O & . & .                 &  .885 & .656      & .291      & .330 \\
                               & . & O & .                 &  .879 & .667      & .114      & .098 \\
                               & O & O & .                 &  .877 & .689      & .067      & .059 \\                     
                               & . & . & O                 &  .874 & .756      & .310      & .212 \\
                               & . & O & O                 &  .866 & .814      & .185      & .065  \\                     
                               & O & O & O                 &  .855 & \bf{\underline{.912}}      & \underline{.055}      & \underline{.030} \\
\hline
\end{tabular}
\end{center}
\caption{\label{table:mitigate-result} Results of bias mitigation methods on \texttt{st} dataset. `O' indicates that the corresponding method is applied. Note that these methods reduce the bias the most for CNN and $\alpha$-GRU when applied together at the same time.}
\end{table}

Table \ref{table:mitigate-result} shows the results of experiments using the three methods proposed. The first rows are the baselines without any method applied. We can see from the second rows of each section that debiased word embeddings alone do not effectively correct the bias of the whole system that well, while gender swapping significantly reduced both the equality difference scores. Meanwhile, fine-tuning bias with a larger, less biased source dataset helped to decrease the equality difference scores and greatly improve the AUC scores from generated unbiased test set. The latter improvement shows that the model significantly reduced errors on the unbiased set in general.

To our surprise, the most effective method was applying both debiased embedding and gender swap to GRU, which reduced the equality differences by 98\% and 89\%, while losing only 1.5\% of the original performance. We assume that this may be related to the influence of ``attending'' model architectures on biases as discussed in Section \ref{sec:measurement-results}. On the other hand, using three methods together improved both generated unbiased set performance and equality differences, but had the largest decrease in the original performance.

All methods involved some performance loss when gender biases were reduced. Especially, fine-tuning had the largest decrease in original test set performance. This could be attributed to the difference in the source and target tasks (abusive/sexist). However, the decrease was marginal (less than 4\%), while the drop in bias was significant. We assume the performance loss happens because mitgation methods modify the data or the model in a way that sometimes deters the models from discriminating important ``unbiased'' features. 

\section{Conclusion}

In this chapter, we worked on abusive language detection using deep learning models. We experimented with many different CNN-based models and showed that using hybrid representations of both characters and words is effective. Moreover, we proposed a two-step approach to divide the problem into detecting abusive language first and then classifying the specific type like sexist or racist afterwards. We believe that two-step approach has potential in that large abusive language datasets with specific label such as profanity, sexist, racist, homophobic, etc. is more difficult to acquire than those simply flagged as abusive. These results related to two-step classification have been published at a workshop addressing abusive language \cite{park2017one}. 

In addition, we discussed model biases, especially toward gender identity terms, in abusive language detection models. We found out that pretrained word embeddings, model architecture, and different datasets all can have influence. Also, we found our proposed methods can reduce gender biases up to 90-98\%, improving the robustness of the models' representations. Although this work focuses on gender terms, the measurement and mitigation methods we proposed can easily be extended to other identity terms on different tasks like sentiment analysis by following similar steps.

As shown in Section \ref{sec:analysis}, some classification performance drop happens when mitigation methods. We believe that a meaningful extension of our work can be developing bias mitigation methods that maintain (or even increase) the classification performance and reduce the bias at the same time. Some previous works \cite{beutel2017data,zhang2018mitigating} employ adversarial training methods to make the classifiers unbiased toward certain variables. However, those works do not deal with natural language where features like gender and race are hidden inside the language. Although those approaches are not directly comparable to our methods, it would be interesting to explore adversarial training in the abusive language detection models in the future. These results related to reducing gender bias have been published at a conference \cite{park2018reducing}. 
 	

%% file: chapter/sec-conclusion.tex
\chapter{Conclusion}\label{sec-conclusion}

In this thesis, we discuss many different representation learning methods in natural language processing problems related to emotions in texts. We use machine learning techniques, especially deep learning models, to improve existing textual representations by accounting for emotions inside text from huge corpora and correcting the biases inside those representations. 

Firstly, we focused on modeling emotions inside a word by training a CNN model to learn each word's emotional word vector (EVEC). Similar to how widely used word embeddings \cite{mikolov2013distributed, pennington2014glove} is used, EVEC can act as a useful input representation for tasks like sentiment/emotion analysis. We conducted experiments to show that EVEC represent emotions better than other word embeddings and can be easily combined with them for better performance in text classification tasks. Our learned representations can be expressed with fixed-size, real-valued N-dimensional vectors, which make it easy to use for downstream tasks. 

Secondly, we extended our research by learning sentence-level representations of emotions. We used a huge corpus distantly labeled by emojis to train a bidirectional LSTM model that captures emotional similarities among different sentences. Such a model can be an effective feature extractor for many affect-related tasks including sentiment/emotion intensity prediction and multi-label emotion classification. We adopted our proposed sentence-level and word-level representations to participate in a competitive workshop called Semantic Evaluation (SemEval) and achieved to produce a system that can rank top 3 among many systems proposed from the NLP community, proving the effectiveness of our representation learning methods. 

Finally, we tackled automatic abusive language detection, which is a more practical application of representation learning and classification related to emotion. We explored different CNN models that can effectively learn from both character and word level representations, showing the significance of using hybrid representations. Also, we proposed a new approach of taking a step-by-step approach in classification. Moreover, we turn our attention to model bias in our representations, specific to gender identity terms, in abusive language detection models. We introduced how to measure those biases and proposed methods to reduce them by 90-98\% so that representations in abusive language detection models can be more robust and generalizable for practical use.

The main contribution of this thesis is that it emphasizes that good representations of emotions is very important and proposes various methodologies to enhance existing representations for relevant text classification tasks like sentiment/emotion analysis and abusive language detection.  The idea of improving representations in different directions, like taking emotions into account and correcting model biases, is unfamiliar but can be crucial in solving those tasks. We need to be aware of the choices of what kind of models and datasets since they can greatly affect the representations produced for the task. We hope in the future these methods can be further explored in more complicated tasks such as dialogue systems and question-answering.

%% file: chapter/sec-publication.tex
\chapter*{Publications}

\begin{enumerate}

\item Ji Ho Park, Jamin Shin, and Pascale Fung. Reducing Gender Bias in Abusive Language Detection. \textit{Proceedings of the 2018 Conference on Empirical Methods in Natural Language Processing (EMNLP)}, 2018.

\item Ji Ho Park, Peng Xu, and Pascale Fung. Plusemo2vec at Semeval-2018 task 1: Exploiting emotion knowledge from emoji and \#hashtags. \textit{Proceedings of The 12th International Workshop on Semantic Evaluation, NAACL-HLT}, 2018.
\item Ji Ho Park and Pascale Fung. One-step and two-step classification for abusive language detection on twitter. \textit{ALW1: 1st Workshop on Abusive Language Online, Annual meeting of the Association of Computational Linguistics (ACL)}, 2017.
\item Ji Ho Park, Nayeon Lee, Dario Bertero, Anik Dey, and Pascale Fung. Emojive! col- lecting emotion data from speech and facial expression using mobile game app. \textit{Interspeech 2017, Show-and-tell Session}, 2017.
\end{enumerate}

%% file: main.bbl
\begin{thebibliography}{10}

\bibitem{badjatiya2017deep}
Pinkesh Badjatiya, Shashank Gupta, Manish Gupta, and Vasudeva Varma.
\newblock Deep learning for hate speech detection in tweets.
\newblock In {\em Proceedings of the 26th International Conference on World
  Wide Web Companion}, pages 759--760. International World Wide Web Conferences
  Steering Committee, 2017.

\bibitem{balikas2016twise}
Georgios Balikas and Massih-Reza Amini.
\newblock Twise at semeval-2016 task 4: Twitter sentiment classification.
\newblock {\em Proceedings of the 10th International Workshop on Semantic
  Evaluation (SemEval-2016)}, 2016.

\bibitem{barrett2006solving}
Lisa~Feldman Barrett.
\newblock Solving the emotion paradox: Categorization and the experience of
  emotion.
\newblock {\em Personality and social psychology review}, 10(1):20--46, 2006.

\bibitem{bengio2013representation}
Yoshua Bengio, Aaron Courville, and Pascal Vincent.
\newblock Representation learning: A review and new perspectives.
\newblock {\em IEEE transactions on pattern analysis and machine intelligence},
  35(8):1798--1828, 2013.

\bibitem{beutel2017data}
Alex Beutel, Jilin Chen, Zhe Zhao, and Ed~H Chi.
\newblock Data decisions and theoretical implications when adversarially
  learning fair representations.
\newblock {\em FAT/ML 2018: 5th Workshop on Fairness, Accountability, and
  Transparency in Machine Learning}.

\bibitem{bojanowski2016enriching}
Piotr Bojanowski, Edouard Grave, Armand Joulin, and Tomas Mikolov.
\newblock Enriching word vectors with subword information.
\newblock {\em Transactions of the Association for Computational Linguistics
  – Volume 5, Issue 1}, 2017.

\bibitem{bolukbasi2016man}
Tolga Bolukbasi, Kai-Wei Chang, James~Y Zou, Venkatesh Saligrama, and Adam~T
  Kalai.
\newblock Man is to computer programmer as woman is to homemaker? debiasing
  word embeddings.
\newblock In {\em Advances in Neural Information Processing Systems}, pages
  4349--4357, 2016.

\bibitem{caliskan2017semantics}
Aylin Caliskan, Joanna~J Bryson, and Arvind Narayanan.
\newblock Semantics derived automatically from language corpora contain
  human-like biases.
\newblock {\em Science}, 356(6334):183--186, 2017.

\bibitem{cho2014learning}
Kyunghyun Cho, Bart Van~Merri{\"e}nboer, Caglar Gulcehre, Dzmitry Bahdanau,
  Fethi Bougares, Holger Schwenk, and Yoshua Bengio.
\newblock Learning phrase representations using rnn encoder-decoder for
  statistical machine translation.
\newblock {\em EMNLP2014}, 2014.

\bibitem{Chollet2015}
Fran\c{c}ois Chollet et~al.
\newblock Keras.
\newblock \url{https://github.com/fchollet/keras}, 2015.

\bibitem{coates2011analysis}
Adam Coates, Andrew Ng, and Honglak Lee.
\newblock An analysis of single-layer networks in unsupervised feature
  learning.
\newblock In {\em Proceedings of the fourteenth international conference on
  artificial intelligence and statistics}, pages 215--223, 2011.

\bibitem{Collobert2011}
Ronan Collobert, Jason Weston, Léon Bottou, Michael Karlen, Koray Kavukcuoglu,
  and Pavel Kuksa.
\newblock Natural language processing (almost) from scratch.
\newblock {\em Journal of Machine Learning Research}, 12(Aug):2493--2537, 2011.

\bibitem{damasio1998emotion}
Antonio~R Damasio.
\newblock Emotion in the perspective of an integrated nervous system1.
\newblock {\em Brain research reviews}, 26(2-3):83--86, 1998.

\bibitem{davidson2017hateoffensive}
Thomas Davidson, Dana Warmsley, Michael Macy, and Ingmar Weber.
\newblock Automated hate speech detection and the problem of offensive
  language.
\newblock In {\em Proceedings of the 11th International AAAI Conference on Web
  and Social Media}, ICWSM '17, pages 512--515, 2017.

\bibitem{Kingma2015}
Jimmy~Ba Diederik P.~Kingma.
\newblock Adam: A method for stochastic optimization.
\newblock In {\em The International Conference on Learning Representations},
  2015.

\bibitem{dixon2017measuring}
Lucas Dixon, John Li, Jeffrey Sorensen, Nithum Thain, and Lucy Vasserman.
\newblock Measuring and mitigating unintended bias in text classification.
\newblock In {\em AAAI}, 2017.

\bibitem{djuric2015hate}
Nemanja Djuric, Jing Zhou, Robin Morris, Mihajlo Grbovic, Vladan Radosavljevic,
  and Narayan Bhamidipati.
\newblock Hate speech detection with comment embeddings.
\newblock In {\em Proceedings of the 24th international conference on world
  wide web}, pages 29--30. ACM, 2015.

\bibitem{ekman1992argument}
Paul Ekman.
\newblock An argument for basic emotions.
\newblock {\em Cognition \& emotion}, 6(3-4):169--200, 1992.

\bibitem{ekman1993facial}
Paul Ekman.
\newblock Facial expression and emotion.
\newblock {\em American psychologist}, 48(4):384, 1993.

\bibitem{felbo2017using}
Bjarke Felbo, Alan Mislove, Anders S{\o}gaard, Iyad Rahwan, and Sune Lehmann.
\newblock Using millions of emoji occurrences to learn any-domain
  representations for detecting sentiment, emotion and sarcasm.
\newblock {\em Conference on Empirical Methods in Natural Language
  Processing(EMNLP) 2017}, 2017.

\bibitem{founta2018large}
Antigoni-Maria Founta, Constantinos Djouvas, Despoina Chatzakou, Ilias
  Leontiadis, Jeremy Blackburn, Gianluca Stringhini, Athena Vakali, Michael
  Sirivianos, and Nicolas Kourtellis.
\newblock Large scale crowdsourcing and characterization of twitter abusive
  behavior.
\newblock {\em AAAI}, 2018.

\bibitem{fung2015robots}
Pascale Fung.
\newblock Robots with heart.
\newblock {\em Scientific American}, 313(5):60--63, 2015.

\bibitem{grone1990laplacian}
Robert Grone, Russell Merris, and V~S\_ Sunder.
\newblock The laplacian spectrum of a graph.
\newblock {\em SIAM Journal on Matrix Analysis and Applications},
  11(2):218--238, 1990.

\bibitem{hill2016learning}
Felix Hill, Kyunghyun Cho, and Anna Korhonen.
\newblock Learning distributed representations of sentences from unlabelled
  data.
\newblock {\em Proceedings of NAACL-HLT 2016}, 2016.

\bibitem{hosseini2017deceiving}
Hossein Hosseini, Sreeram Kannan, Baosen Zhang, and Radha Poovendran.
\newblock Deceiving google's perspective api built for detecting toxic
  comments.
\newblock {\em In Proceedings of the Workshop on Natural Language Processing
  for ComputerMediated Communication (NLP4CMC)}, 2017.

\bibitem{iyyer2015deep}
Mohit Iyyer, Varun Manjunatha, Jordan Boyd-Graber, and Hal Daum{\'e}~III.
\newblock Deep unordered composition rivals syntactic methods for text
  classification.
\newblock In {\em Proceedings of the 53rd Annual Meeting of the Association for
  Computational Linguistics and the 7th International Joint Conference on
  Natural Language Processing (Volume 1: Long Papers)}, volume~1, pages
  1681--1691, 2015.

\bibitem{joulin2016bag}
Armand Joulin, Edouard Grave, Piotr Bojanowski, and Tomas Mikolov.
\newblock Bag of tricks for efficient text classification.
\newblock {\em Proceedings of the 15th Conference of the European Chapter of
  the Association for Computational Linguistics: Volume 2, Short Paper}, 2016.

\bibitem{kalchbrennerconvolutional}
Nal Kalchbrenner, Edward Grefenstette, and Phil Blunsom.
\newblock A convolutional neural network for modelling sentences.

\bibitem{Kim2014}
Yoon Kim.
\newblock Convolutional neural networks for sentence classification.
\newblock In {\em Conference on Empirical Methods in Natural Language
  Processing(EMNLP), publisher={Citeseer},}, 2014.

\bibitem{kiritchenko2018examining}
Svetlana Kiritchenko and Saif~M Mohammad.
\newblock Examining gender and race bias in two hundred sentiment analysis
  systems.
\newblock {\em Proceedings of the 7th Joint Conference on Lexical and
  Computational Semantics(*SEM), New Orleans, USA}, 2018.

\bibitem{kiros2015skip}
Ryan Kiros, Yukun Zhu, Ruslan~R Salakhutdinov, Richard Zemel, Raquel Urtasun,
  Antonio Torralba, and Sanja Fidler.
\newblock Skip-thought vectors.
\newblock In {\em Advances in neural information processing systems}, pages
  3294--3302, 2015.

\bibitem{le2014distributed}
Quoc Le and Tomas Mikolov.
\newblock Distributed representations of sentences and documents.
\newblock In {\em International Conference on Machine Learning}, pages
  1188--1196, 2014.

\bibitem{Lecun2015}
Yann LeCun, Yoshua Bengio, and Geoffrey Hinton.
\newblock Deep learning.
\newblock {\em Nature}, 521(7553):436--444, 2015.

\bibitem{Li2016}
Jiwei Li, Xinlei Chen, Eduard Hovy, and Dan Jurafsky.
\newblock Visualizing and understanding neural models in nlp.
\newblock In {\em Proceedings of NAACL-HLT}, pages 681--691, 2016.

\bibitem{liang2005semi}
Percy Liang.
\newblock {\em Semi-supervised learning for natural language}.
\newblock PhD thesis, Massachusetts Institute of Technology, 2005.

\bibitem{lindquist2015role}
Kristen~A Lindquist, Jennifer~K MacCormack, and Holly Shablack.
\newblock The role of language in emotion: predictions from psychological
  constructionism.
\newblock {\em Frontiers in psychology}, 6:444, 2015.

\bibitem{liu2012survey}
Bing Liu and Lei Zhang.
\newblock A survey of opinion mining and sentiment analysis.
\newblock In {\em Mining text data}, pages 415--463. Springer, 2012.

\bibitem{meyer}
Robinson Meyer.
\newblock Twitter's famous racist problem.
\newblock {\em The Atlantic}, 07/21 2016.

\bibitem{mikolov2013distributed}
Tomas Mikolov, Ilya Sutskever, Kai Chen, Greg~S Corrado, and Jeff Dean.
\newblock Distributed representations of words and phrases and their
  compositionality.
\newblock In {\em Advances in neural information processing systems}, pages
  3111--3119, 2013.

\bibitem{Mohammad2017}
Saif Mohammad and Felipe Bravo-Marquez.
\newblock Wassa-2017 shared task on emotion intensity.
\newblock In {\em Proceedings of the 8th Workshop on Computational Approaches
  to Subjectivity, Sentiment and Social Media Analysis}, pages 34--49, 2017.

\bibitem{mohammad2018semeval}
Saif~M Mohammad, Felipe Bravo-Marquez, Mohammad Salameh, and Svetlana
  Kiritchenko.
\newblock Semeval-2018 task 1: Affect in tweets.
\newblock In {\em Proceedings of International Workshop on Semantic Evaluation
  (SemEval-2018), New Orleans, LA, USA}, 2018.

\bibitem{LREC18-TweetEmo}
Saif~M. Mohammad and Svetlana Kiritchenko.
\newblock Understanding emotions: A dataset of tweets to study interactions
  between affect categories.
\newblock In {\em Proceedings of the 11th Edition of the Language Resources and
  Evaluation Conference}, Miyazaki, Japan, 2018.

\bibitem{Mohammad2013}
Saif~M. Mohammad and Peter~D. Turney.
\newblock Crowdsourcing a word–emotion association lexicon.
\newblock {\em Computational Intelligence}, 29(3):436--465, 2013.

\bibitem{Nair2010}
Vinod Nair and Geoffrey~E Hinton.
\newblock Rectified linear units improve restricted boltzmann machines.
\newblock In {\em Proceedings of the 27th international conference on machine
  learning (ICML-10)}, pages 807--814, 2010.

\bibitem{nobata2016abusive}
Chikashi Nobata, Joel Tetreault, Achint Thomas, Yashar Mehdad, and Yi~Chang.
\newblock Abusive language detection in online user content.
\newblock In {\em Proceedings of the 25th international conference on world
  wide web}, pages 145--153. International World Wide Web Conferences Steering
  Committee, 2016.

\bibitem{panksepp2004affective}
Jaak Panksepp.
\newblock {\em Affective neuroscience: The foundations of human and animal
  emotions}.
\newblock Oxford university press, 2004.

\bibitem{papegnies2017detection}
Etienne Papegnies, Vincent Labatut, Richard Dufour, and Georges Linar{\`e}s.
\newblock Detection of abusive messages in an on-line community.
\newblock In {\em 14{\`e}me Conf{\'e}rence en Recherche d'Information et
  Applications (CORIA)}, pages 153--168, 2017.

\bibitem{park2018reducing}
Jamin~Shin Park, Ji~Ho and Pascale Fung.
\newblock Reducing gender bias in abusive language detection.
\newblock {\em Conference on Empirical Methods in Natural Language
  Processing(EMNLP)}, 2018.

\bibitem{park2017one}
Ji~Ho Park and Pascale Fung.
\newblock One-step and two-step classification for abusive language detection
  on twitter.
\newblock {\em ALW1: 1st Workshop on Abusive Language Online, Annual meeting of
  the Association of Computational Linguistics (ACL)}, 2017.

\bibitem{park2018plusemo2vec}
Ji~Ho Park, Peng Xu, and Pascale Fung.
\newblock Plusemo2vec at semeval-2018 task 1: Exploiting emotion knowledge from
  emoji and\# hashtags.
\newblock {\em Proceedings of The 12th International Workshop on Semantic
  Evaluation, NAACL-HLT 2018}, 2018.

\bibitem{paszke2017automatic}
Adam Paszke, Sam Gross, Soumith Chintala, Gregory Chanan, Edward Yang, Zachary
  DeVito, Zeming Lin, Alban Desmaison, Luca Antiga, and Adam Lerer.
\newblock Automatic differentiation in pytorch.
\newblock 2017.

\bibitem{pavlopoulos2017deeper}
John Pavlopoulos, Prodromos Malakasiotis, and Ion Androutsopoulos.
\newblock Deeper attention to abusive user content moderation.
\newblock In {\em Proceedings of the 2017 Conference on Empirical Methods in
  Natural Language Processing}, pages 1125--1135, 2017.

\bibitem{pedregosa2011scikit}
Fabian Pedregosa, Ga{\"e}l Varoquaux, Alexandre Gramfort, Vincent Michel,
  Bertrand Thirion, Olivier Grisel, Mathieu Blondel, Peter Prettenhofer, Ron
  Weiss, Vincent Dubourg, et~al.
\newblock Scikit-learn: Machine learning in python.
\newblock {\em Journal of machine learning research}, 12(Oct):2825--2830, 2011.

\bibitem{pennington2014glove}
Jeffrey Pennington, Richard Socher, and Christopher Manning.
\newblock Glove: Global vectors for word representation.
\newblock In {\em Proceedings of the 2014 conference on empirical methods in
  natural language processing (EMNLP)}, pages 1532--1543, 2014.

\bibitem{perse}
Elizabeth~M. Perse and Jennifer Lambe.
\newblock {\em Media effects and society}.
\newblock Routledge, 2016.

\bibitem{plutchik1984emotions}
Robert Plutchik.
\newblock Emotions: A general psychoevolutionary theory.
\newblock {\em Approaches to emotion}, 1984:197--219, 1984.

\bibitem{read2009classifier}
Jesse Read, Bernhard Pfahringer, Geoff Holmes, and Eibe Frank.
\newblock Classifier chains for multi-label classification.
\newblock In {\em Joint European Conference on Machine Learning and Knowledge
  Discovery in Databases}, pages 254--269. Springer, 2009.

\bibitem{ross2017measuring}
Bj{\"o}rn Ross, Michael Rist, Guillermo Carbonell, Benjamin Cabrera, Nils
  Kurowsky, and Michael Wojatzki.
\newblock Measuring the reliability of hate speech annotations: The case of the
  european refugee crisis.
\newblock {\em arXiv preprint arXiv:1701.08118}, 2017.

\bibitem{ross}
Björn Ross, Michael Rist, Guillermo Carbonell, Benjamin Cabrera, Nils
  Kurowsky, and Michael Wojatzki.
\newblock Measuring the reliability of hate speech annotations: The case of the
  european refugee crisis.
\newblock {\em arXiv preprint arXiv:1701.08118}, 2017.

\bibitem{schmidt}
Anna Schmidt and Michael Wiegand.
\newblock A survey on hate speech detection using natural language processing.
\newblock {\em SocialNLP 2017}, page~1, 2017.

\bibitem{schnabel2015evaluation}
Tobias Schnabel, Igor Labutov, David Mimno, and Thorsten Joachims.
\newblock Evaluation methods for unsupervised word embeddings.
\newblock In {\em Proceedings of the 2015 Conference on Empirical Methods in
  Natural Language Processing}, pages 298--307, 2015.

\bibitem{schuster1997bidirectional}
Mike Schuster and Kuldip~K Paliwal.
\newblock Bidirectional recurrent neural networks.
\newblock {\em IEEE Transactions on Signal Processing}, 45(11):2673--2681,
  1997.

\bibitem{shahid2016pca}
Nauman Shahid, Nathanael Perraudin, Vassilis Kalofolias, Benjamin Ricaud, and
  Pierre Vandergheynst.
\newblock Pca using graph total variation.
\newblock In {\em Acoustics, Speech and Signal Processing (ICASSP), 2016 IEEE
  International Conference on}, pages 4668--4672. Ieee, 2016.

\bibitem{Shaver1987}
Phillip Shaver, Judith Schwartz, Donald Kirson, and Cary O'connor.
\newblock Emotion knowledge: Further exploration of a prototype approach.
\newblock {\em Journal of personality and social psychology}, 52(6):1061, 1987.

\bibitem{Simonyan2013}
Karen Simonyan, Andrea Vedaldi, and Andrew Zisserman.
\newblock Deep inside convolutional networks: Visualising image classification
  models and saliency maps.
\newblock In {\em Workshop of International Conference on Learning
  Representations(ICLR)}, 2014.

\bibitem{socher2013recursive}
Richard Socher, Alex Perelygin, Jean Wu, Jason Chuang, Christopher~D Manning,
  Andrew Ng, and Christopher Potts.
\newblock Recursive deep models for semantic compositionality over a sentiment
  treebank.
\newblock In {\em Proceedings of the 2013 conference on empirical methods in
  natural language processing}, pages 1631--1642, 2013.

\bibitem{suttles2013distant}
Jared Suttles and Nancy Ide.
\newblock Distant supervision for emotion classification with discrete binary
  values.
\newblock In {\em International Conference on Intelligent Text Processing and
  Computational Linguistics}, pages 121--136. Springer, 2013.

\bibitem{tang2014building}
Duyu Tang, Furu Wei, Bing Qin, Ming Zhou, and Ting Liu.
\newblock Building large-scale twitter-specific sentiment lexicon: A
  representation learning approach.
\newblock In {\em Proceedings of COLING 2014, the 25th International Conference
  on Computational Linguistics: Technical Papers}, pages 172--182, 2014.

\bibitem{tang2014learning}
Duyu Tang, Furu Wei, Nan Yang, Ming Zhou, Ting Liu, and Bing Qin.
\newblock Learning sentiment-specific word embedding for twitter sentiment
  classification.
\newblock In {\em Proceedings of the 52nd Annual Meeting of the Association for
  Computational Linguistics (Volume 1: Long Papers)}, volume~1, pages
  1555--1565, 2014.

\bibitem{Thelwall2012}
Mike Thelwall, Kevan Buckley, and Georgios Paltoglou.
\newblock Sentiment strength detection for the social web.
\newblock {\em Journal of the Association for Information Science and
  Technology}, 63(1):163--173, 2012.

\bibitem{Thelwall2010}
Mike Thelwall, Kevan Buckley, Georgios Paltoglou, Di~Cai, and Arvid Kappas.
\newblock Sentiment strength detection in short informal text.
\newblock {\em Journal of the Association for Information Science and
  Technology}, 61(12):2544--2558, 2010.

\bibitem{Wallbott1986}
Harald~G. Wallbott and Klaus~R. Scherer.
\newblock How universal and specific is emotional experience? evidence from 27
  countries on five continents.
\newblock {\em Information (International Social Science Council)},
  25(4):763--795, 1986.

\bibitem{wang2012harnessing}
Wenbo Wang, Lu~Chen, Krishnaprasad Thirunarayan, and Amit~P Sheth.
\newblock Harnessing twitter" big data" for automatic emotion identification.
\newblock In {\em Privacy, Security, Risk and Trust (PASSAT), 2012
  International Conference on and 2012 International Confernece on Social
  Computing (SocialCom)}, pages 587--592. IEEE, 2012.

\bibitem{Wang2012}
Wenbo Wang, Lu~Chen, Krishnaprasad Thirunarayan, and Amit~P. Sheth.
\newblock Harnessing twitter" big data" for automatic emotion identification.
\newblock In {\em Privacy, Security, Risk and Trust (PASSAT), 2012
  International Conference on and 2012 International Confernece on Social
  Computing (SocialCom)}, pages 587--592. IEEE, 2012.

\bibitem{warner2012detecting}
William Warner and Julia Hirschberg.
\newblock Detecting hate speech on the world wide web.
\newblock In {\em Proceedings of the Second Workshop on Language in Social
  Media}, pages 19--26. Association for Computational Linguistics, 2012.

\bibitem{wassem}
Zeerak Waseem.
\newblock Are you a racist or am i seeing things? annotator influence on hate
  speech detection on twitter.
\newblock In {\em Proceedings of the 1st Workshop on Natural Language
  Processing and Computational Social Science}, pages 138--142, 2016.

\bibitem{waseemhovy}
Zeerak Waseem and Dirk Hovy.
\newblock Hateful symbols or hateful people? predictive features for hate
  speech detection on twitter.
\newblock In {\em Proceedings of NAACL-HLT}, pages 88--93, 2016.

\bibitem{wood2016ruder}
Ian Wood and Sebastian Ruder.
\newblock Emoji as emotion tags for tweets.
\newblock In {\em Emotion and Sentiment Analysis Workshop, at LREC2016}.
  LREC2016, 2016.

\bibitem{wulczyn2017ex}
Ellery Wulczyn, Nithum Thain, and Lucas Dixon.
\newblock Ex machina: Personal attacks seen at scale.
\newblock In {\em Proceedings of the 26th International Conference on World
  Wide Web}, pages 1391--1399. International World Wide Web Conferences
  Steering Committee, 2017.

\bibitem{yang2016hierarchical}
Zichao Yang, Diyi Yang, Chris Dyer, Xiaodong He, Alex Smola, and Eduard Hovy.
\newblock Hierarchical attention networks for document classification.
\newblock In {\em Proceedings of the 2016 Conference of the North American
  Chapter of the Association for Computational Linguistics: Human Language
  Technologies}, pages 1480--1489, 2016.

\bibitem{yu2017refining}
Liang-Chih Yu, Jin Wang, K~Robert Lai, and Xuejie Zhang.
\newblock Refining word embeddings for sentiment analysis.
\newblock In {\em Proceedings of the 2017 Conference on Empirical Methods in
  Natural Language Processing}, pages 534--539, 2017.

\bibitem{zhang2018mitigating}
Brian~Hu Zhang, Blake Lemoine, and Margaret Mitchell.
\newblock Mitigating unwanted biases with adversarial learning.
\newblock {\em Proceedings of AAAI/ACM Conference on Ethics and Society(AIES)
  2018}, 2018.

\bibitem{zhang2015character}
Xiang Zhang, Junbo Zhao, and Yann LeCun.
\newblock Character-level convolutional networks for text classification.
\newblock In {\em Advances in neural information processing systems}, pages
  649--657, 2015.

\bibitem{zhao2018gender}
Jieyu Zhao, Tianlu Wang, Mark Yatskar, Vicente Ordonez, and Kai-Wei Chang.
\newblock Gender bias in coreference resolution: Evaluation and debiasing
  methods.
\newblock {\em NAACL 2018}, 2018.

\bibitem{zhou2016ecnu}
Yunxiao Zhou, Zhihua Zhang, and Man Lan.
\newblock Ecnu at semeval-2016 task 4: An empirical investigation of
  traditional nlp features and word embedding features for sentence-level and
  topic-level sentiment analysis in twitter.
\newblock In {\em Proceedings of the 10th International Workshop on Semantic
  Evaluation (SemEval-2016)}, pages 256--261, 2016.

\end{thebibliography}
